\RequirePackage{rotating} 
\documentclass[11pt]{article}

\usepackage{boxedminipage}
\usepackage[vlined, ruled]{algorithm2e}
\newcommand{\argmin}{\operatornamewithlimits{arg\,min}}
\usepackage{titlesec}
\usepackage[utf8]{inputenc}
\usepackage{caption, makecell}

\usepackage{threeparttable}

\usepackage{geometry}
\usepackage{mdframed}
\usepackage{booktabs}

\titleformat{\paragraph}
{\normalfont\normalsize\itshape}{\theparagraph}{1em}{}
\titlespacing*{\paragraph}
{0pt}{3.25ex plus 1ex minus .2ex}{1.5ex plus .2ex}

\usepackage{graphicx}
\usepackage{mathrsfs,lscape,subfigure,enumerate,epstopdf}
\usepackage{amssymb}
\usepackage{pifont}
\usepackage{amsmath,lipsum}
\usepackage{amsthm}
\usepackage{enumerate} 
\usepackage{pgfplots}
\usepackage{tikz}
\usepackage{bm}
\usepackage{float}

\usepackage{glossaries}
\pgfplotsset{width=10cm,compat=1.16}
\newacronym{bss}{BSS}{Bike-sharing systems}
\newacronym{sbrp}{SBRP}{Static Bicycle Repositioning Problem}
\newacronym{dbrp}{DBRP}{Dynamic Bicycle Repositioning Problem}
\newacronym{gbfs}{GBFS}{General Bikeshare Feed Specification}

\newacronym{mdp}{MDP}{Markov Decision Process}
\newacronym{mmdp}{MMDP}{Multi-agent Markov Decision Process}
\newacronym{mas}{MAS}{Multi-agent System}
\newacronym{nn}{NN}{Neural Networks}
\newacronym{rf}{RF}{random forest}
\newacronym{rl}{RL}{Reinforcement Learning}
\newacronym{drl}{DRL}{Deep Reinforcement Learning}
\newacronym{ml}{ML}{Machine Learning}
\newacronym{pfa}{PFA}{Policy Function Approximation}
\newacronym{dqn}{DQN}{Deep Q-Network}
\newacronym{ddqn}{DDQN}{Double Deep Q-Network}
\newacronym{vdn}{VDN}{Value-decomposition networks} 
\newacronym{td}{TD}{Temporal-Difference}
\newacronym{ctde}{CTDE}{Centralized Training Decentralized Execution}
\newacronym{marl}{MARL}{Multi-Agent Reinforcement Learning}
\newacronym{tsp}{TSP}{Traveling Salesman Problem}
\newacronym{lp}{LP}{Linear Programming}
\newacronym{ip}{IP}{Integer Programming}
\newacronym{milp}{MILP}{Mixed-Integer Linear Programming}
\newacronym{mip}{MIP}{Mixed-Integer Programming}
\newacronym{qp}{QP}{quadratic programming}
\newacronym{miqp}{MIQP}{mixed-integer quadratic programming}
\newacronym{pdp}{PDP}{Pickup and Delivery Problem}
\newacronym{vrp}{VRP}{Vehicle Routing Problem}
\newacronym{vrptw}{VRPTW}{Vehicle Routing Problem with Time Windows}
\newacronym{vns}{VNS}{Variable Neighborhood Search}
\newacronym{cp}{CP}{Constraint Programming}
\newacronym{ilp}{ILP}{Integer Linear Programming} 
\newacronym{aco}{ACO}{Ant Colony Optimization}  
\newacronym{ga}{GA}{Genetic Algorithm} 

\newacronym{bnb}{B\&B}{branch-and-bound}
\newacronym{bnc}{B\&C}{branch-and-cut}
\newacronym{lpp}{LPP}{load planning problem}
\newacronym{lb}{LB}{local branching}
\newacronym{lbc}{LBCs}{local branching cuts}
\newacronym{gap}{GAP}{generalized assignment problem}
\newacronym{mvc}{MVC}{minimum vertex cover}
\newacronym{mc}{MC}{maximum cut}

\usetikzlibrary{shapes.geometric, arrows, calc, fit, positioning, chains, arrows.meta,decorations.pathreplacing,matrix,backgrounds,positioning,shapes}

\glsdisablehyper 
\makeglossaries

\usepackage{lineno}
\usepackage{longtable}
\usepackage{threeparttable}
\usepackage{multicol}
\usepackage{multirow}
\usepackage{booktabs}
\usepackage{diagbox}
\usepackage{natbib}
\setcitestyle{square,numbers}
\usepackage{fancybox}
\usepackage{textcomp}
\usepackage{rotating}
\usepackage{pdflscape} 

\usepackage{colortbl}
\usepackage{tabulary}
\usepackage{etoolbox}
\usetikzlibrary{patterns}
\usepackage{afterpage,lscape}
\usetikzlibrary{decorations.pathreplacing,calligraphy}
\usepackage[colorlinks]{hyperref}
\usepackage{authblk}

\usepackage{mathrsfs}

\usepackage[titletoc,title]{appendix}

\usepackage{algpseudocode}
\usepackage{tikz}
\usepackage{pgfplots}
\usepackage{setspace}
\usepackage{etoolbox}
\usepackage[vlined, ruled]{algorithm2e}
\usepackage{amsmath,amssymb,amsfonts}
\usetikzlibrary{positioning, fit, backgrounds, shadows,arrows.meta}
\usetikzlibrary{shapes.geometric, arrows, calc, fit, positioning, chains, arrows.meta,decorations.pathreplacing,matrix,backgrounds, shapes, calligraphy}
\usetikzlibrary{arrows.meta}
\usepackage{fontawesome5}
\newcommand\bike{\faBicycle}
\newcommand\dock{\faParking}

\title{Single- vs. Dual-Policy Reinforcement Learning for Dynamic Bike Rebalancing}

\author[1]{Jiaqi Liang}

\author[1]{Defeng Liu}

\author[2]{Sanjay Dominik Jena}

\author[3]{Andrea Lodi}

\author[1]{Thibaut Vidal}

\affil[1]{\itshape\small Department of Mathematical and Industrial Engineering, Polytechnique Montréal, 2500 Chemin de Polytechnique, Montréal, H3T1J4, Canada}

\affil[2]{\itshape\small School of Management, Université du Québec à Montréal, 315 Rue Sainte-Catherine Est, Montréal,  H2X3X2, Canada}

\affil[3]{\itshape\small Cornell Tech and Technion - IIT, 2 West Loop Road,
New York, 10044, USA}

\date{\vspace{-5ex}}
\providecommand{\keywords}[1]{\textbf{\textit{Keywords:}} #1}

\begin{document}
\maketitle

\begin{abstract}
Bike-sharing systems (BSS) provide a sustainable urban mobility solution, but ensuring their reliability requires effective rebalancing strategies to address stochastic demand and prevent station imbalances. This paper proposes reinforcement learning (RL) algorithms for dynamic rebalancing problem with multiple vehicles, introducing and comparing two RL approaches: Single-policy RL and Dual-policy RL. We formulate this network optimization problem as a Markov Decision Process within a continuous-time framework, allowing vehicles to make independent and cooperative rebalancing decisions without synchronization constraints. In the first approach, a single deep Q-network (DQN) is trained to jointly learn inventory and routing decisions. The second approach decouples node-level inventory decisions from arc-level vehicle routing, enhancing learning efficiency and adaptability. A high-fidelity simulator under the first-arrive-first-serve rule is developed to estimate rewards across diverse demand scenarios influenced by temporal and weather variations. Extensive experiments demonstrate that while the single-policy model is competitive against several benchmarks, the dual-policy model significantly reduces lost demand. These findings provide valuable insights for bike-sharing operators, reinforcing the potential of RL for real-time rebalancing and paving the way for more adaptive and intelligent urban mobility solutions.
\end{abstract}


\section{Introduction}\label{sec:Intro}

The development of Bike-Sharing Systems (BSS) has been instrumental in alleviating urban traffic congestion and reducing CO$_2$ emissions \citep{meddin2022}. However, the stochastic nature of user behavior and peak hour demands often lead to unbalanced stations, necessitating efficient rebalancing strategies to minimize lost demand.

A BSS can be represented by a directed graph: stations are nodes with limited docking capacity, and routes between stations are arcs with travel times. The Dynamic Bike Repositioning Problem (DBRP) addresses the real-time redistribution of bikes across a network of stations to meet fluctuating demand during the day \citep{raviv2013static}. Typically, operators employ a fleet of vehicles to redistribute bikes between stations, aiming to minimize unmet demand. This task involves making both inventory decisions (how many bikes to pick up or drop off at each node) and routing decisions (which station to visit next). The left panel of Figure~\ref{fig:Bss} illustrates a simple BSS example with four stations and a fleet of two vehicles. Each station is annotated with its current bike inventory and the number of available docks. The rental and return events are displayed in the bottom panel as chronological markers. This demand interacts with the system by influencing station inventory levels and guiding vehicle rebalancing decisions. The model considers these interactions to generate optimized rebalancing actions.

Existing works primarily rely on Mixed Integer Programming (MIP) models with time discretization \citep[]{ghosh2019improving, liang2024dynamic}, but must strike a delicate balance between discretization accuracy and computational efficiency. Markov Decision Processes (MDPs) offer an alternative by modeling DBRP as a sequential decision-making problem~\citep[see, e.g.,][]{brinkmann2019dynamic, luo2022dynamic, seo2020dynamic}. Although several studies applied dynamic programming or heuristic policies to tackle the MDP \citep[see, e.g.,][]{legros2019dynamic, brinkmann2019dynamic, brinkmann2020multi}, these approaches rely on having a complete and accurate model of the environment and can face challenges as the network grows. Reinforcement learning (RL), on the other hand, learns policies directly from interactions with the environment, obviating the need for an explicit model of the environment dynamics. This makes RL applicable to scenarios where the system dynamics are complex, unknown, or hard to model accurately. Thus, RL techniques offer the potential to solve MDP models for DBRP, an area that remains underexplored.

Existing MDP models simultaneously make inventory and routing decisions upon a vehicle's arrival at a station with single-vehicle operations \citep[see, e.g.,][]{brinkmann2019dynamic, luo2022dynamic, seo2022rebalancing}, or small networks involving a limited number of stations or service regions \citep[typically 3 to 30, see, e.g.,][]{legros2019dynamic, li2018dynamic, luo2022dynamic}. Moreover, while the inventory decision is being executed, the rest of the network's inventories continue to change, potentially making the routing decision upon the vehicle's arrival suboptimal. Since system dynamics evolve during inventory rebalancing operations, decoupling inventory and routing decisions is critical. To address this, we propose a spatio-temporal RL framework for DBRP under a continuous-time setting, allowing asynchronous and cooperative decision-making among multiple vehicles. Within this framework, we introduce and compare two RL architectures:
\begin{itemize}
    \item Single Policy RL algorithm (SPRL): A Deep Q-Network (DQN) that jointly learns inventory and routing decisions within a single policy. This serves as a strong RL benchmark, aligning with many existing reinforcement learning models in the literature.
    \item Dual Policy RL algorithm (DPRL): A more structured approach where two DQNs are trained separately for inventory decisions (on nodes) and routing decisions (on arcs) while sharing state, as shown in Figure~\ref{fig:Bss}. This decomposition enables a more granular modeling of network dynamics, significantly improving rebalancing efficiency.
\end{itemize}


\begin{figure}[!htbp]
\centering
\resizebox{\textwidth}{!}{\tikzstyle{station} = [
    rectangle,
    draw,
    fill=blue!10,
    text width=4em,
    text centered,
    rounded corners,
    minimum height=4em,
    font=\large
]
\tikzstyle{policy} = [
    rectangle,
    draw,
    fill=green!10,
    text centered,
    rounded corners,
    minimum height=4em,
]
\tikzstyle{policy1} = [
    rectangle,
    draw,
    fill=yellow!20,
    text centered,
    rounded corners,
    minimum height=4em,
]
\tikzstyle{nn} = [
    circle,
    draw,
    fill,
    minimum size=0.3pt,
    scale=0.7,
]
\newcommand\nn{
    \begin{tikzpicture}
        \foreach \i in {1, 2, 3} {
            \node [nn] (Input\i) at (0, \i) {};
        }
        \foreach \h in {1, 2, 3} {
            \node [nn] (Hidden\h) at (1, \h) {};
        }
        \foreach \o in {1, 2} {
            \node [nn] (Output\o) at (2, {\o + 0.5}) {};
        }
        \foreach \i in {1, 2, 3} {
            \foreach \h in {1, 2, 3} {
                \draw (Input\i) -- (Hidden\h);
            }
        }
        \foreach \h in {1, 2, 3} {
            \foreach \o in {1, 2} {
                \draw (Hidden\h) -- (Output\o);
            }
        }
    \end{tikzpicture}
}
\begin{tikzpicture}
\node (StateBefore) {
\begin{tikzpicture}[node distance=3cm]
\node [
    station,
    label=above:Station 1,
] (Station1) {4 \bike \\ 6 \dock};
\node [
    station,
    right of=Station1,
    label=above:Station 2,
] (Station2) {7 \bike \\ 1 \dock};
\node [
    station,
    below of=Station1,
    label=below:Station 3,
] (Station3) {1 \bike \\ 7 \dock};
\node [
    station,
    right of=Station3,
    label=below:Station 4,
] (Station4) {2 \bike \\ 6 \dock};
\node [anchor=west, font={\fontsize{20}{10}\selectfont}] (Truck1) at (Station2.east) {\faTruck};
\node [anchor=east, font={\fontsize{20}{10}\selectfont}] (Truck2) at (Station3.west) {\faTruck};
\node [
    draw,
    rounded corners=5pt,
    line width=1.5pt,
    inner sep=20pt,
    outer sep=0pt,
    fit=(Station1)(Station2)(Station3)(Station4)(Truck1)(Truck2)
] (State) {};
\end{tikzpicture}
};

\node (Policy) [right=of StateBefore, xshift = 1.5cm] {
\begin{tikzpicture}[
    node distance=2cm,
]
\node [
    policy,
] (InventoryPolicy) {\nn};
\node[above=0.5pt of InventoryPolicy] {Inventory Policy};
\node[below=1pt of InventoryPolicy] {Routing Policy};
\node [
    policy1,
    yshift=-19pt,
    anchor=north,
] (RoutingPolicy) at (InventoryPolicy.south) {\nn};

\node [
    draw,
    rounded corners=5pt,
    line width=1.5pt,
    inner sep=20pt,
    outer sep=0pt,
    label=above:Dual Policy,
    fit=(InventoryPolicy)(RoutingPolicy)
] {};
\end{tikzpicture}
};


\node[single arrow, draw=black, fill=white, 
      minimum width = 5mm, single arrow head extend=5pt,
      minimum height=25mm] (ArrowNode) at ($(StateBefore.east)!0.5!(Policy.west)$) {};
\draw [draw=none] (StateBefore.east) -- (Policy.west) node[midway, below=5pt] {State};

\node (StateAfter) [right=of Policy, xshift = 1.5cm] {
\begin{tikzpicture}[node distance=3cm]
\node [
    station,
    label=above:Station 1,
] (Station1) {4 \bike \\ 4 \dock};
\node [
    station,
    right of=Station1,
    label=above:Station 2,
] (Station2) {2 \bike \\ 6 \dock};
\node [
    station,
    below of=Station1,
    label=below:Station 3,
] (Station3) {5 \bike \\ 3 \dock};
\node [
    station,
    right of=Station3,
    label=below:Station 4,
] (Station4) {1 \bike \\ 7 \dock};
\node [anchor=west, font={\fontsize{20}{10}\selectfont}] (Truck1) at (Station2.east) {\faTruck};
\node [anchor=east, font={\fontsize{20}{10}\selectfont}] (Truck2) at (Station3.west) {\faTruck};
\node [
    draw,
    rounded corners=5pt,
    line width=1.5pt,
    inner sep=20pt,
    outer sep=0pt,
    fit=(Station1)(Station2)(Station3)(Station4)(Truck1)(Truck2)
] (State) {};
\node[anchor=north, align=center] at ([xshift=-8pt, yshift=2pt]Truck2.south) {\parbox[c]{2cm}{\centering Drop off\protect\\[-7pt]5 bikes}};
\node[anchor=north, align=center] at ([xshift=8pt, yshift=25pt]Truck1.north) {\parbox[c]{2cm}{\centering Pick up\protect\\[-7pt]6 bikes}};
\draw [ultra thick] (Truck2) edge [->, bend left = 30, >=stealth] (Station1.west);
\draw [ultra thick] (Truck1) edge [->, bend left = 30, >=stealth] (Station4.east);
\draw[->, ultra thick, >=stealth, red] (Station4.north) ++(0.8,-0.3) -- ++(0.0,-0.5);
\draw[->, ultra thick, >=stealth, red] (Station3.north) ++(0.8,-0.3) -- ++(0.0,-0.5);
\draw[->, ultra thick, >=stealth, blue] (Station2.south) ++(0.8,0.3) -- ++(0.0,0.5);
\end{tikzpicture}
};

\node[single arrow, draw=black, fill=white, 
      minimum width = 5mm, single arrow head extend=5pt,
      minimum height=25mm] (ArrowNode) at ($(Policy.east)!0.5!(StateAfter.west)$) {};
\draw [draw=none] (Policy.east) -- (StateAfter.west) node[midway, below=5pt] {\parbox{2cm}{Rebalancing \\ Solution}};

\node (DemandAfter) [below=of Policy]{
\begin{tikzpicture}[node distance=2cm]
\draw[->, ultra thick, >=stealth] (0,0) -- (10,0) node[above] {\Large \(t\)};
\foreach \rental in {1, 3, 4, 6, 7, 8} {
    \draw[->, ultra thick, >=stealth, red] (\rental,1) -- (\rental,0);
}
\foreach \return in {2, 5, 9} {
    \draw[->, ultra thick, >=stealth, blue] (\return,-1) -- (\return,0);
}
\fill[gray, opacity=0.6] (0,-1) rectangle (3.5,1);

\node[left, text=red] at (-0.5,0.5) {Rental};

\node[left, text=blue] at (-0.5,-0.5) {Return};

\node[left] at (-2.5,1) {Demand};

\end{tikzpicture}
};

\node [
    draw,
    rounded corners=5pt,
    line width=1.5pt,
    inner sep=10pt,
    outer sep=0pt,
    fit=(DemandAfter)
] (Demand) {};

\draw [->, ultra thick] (Demand.west) -| (StateBefore.south);
\draw [->, ultra thick] (Demand.east) -| (StateAfter.south);
\end{tikzpicture}}
\caption{Dynamic rebalancing in BSS and DPRL: Inventory information, vehicle location and inventory level, and user demand typically serve as inputs in dynamic rebalancing models. The proposed DPRL generates rebalancing solutions based on the environmental interactions among stations, vehicles, and users.}
\label{fig:Bss}
\end{figure}

To facilitate the learning process, we use a real-time BSS simulator to evaluate immediate rewards based on the BSS environment and various user demand scenarios. We conduct extensive experiments, including comparisons with MIP models and RL architectures, as well as ablation studies. DPRL significantly outperforms all benchmarks, demonstrating its effectiveness for real-time rebalancing applications. Since other related problems, such as pickup-and-delivery and ride-sharing, share many common traits with the DBRP, involving both inventory and routing decisions, we believe that our approach could be adapted to other network and transportation optimization applications beyond BSS.

The remainder of the paper is organized as follows. Section~\ref{sec:literature} reviews related literature. Section~\ref{sec:Model} models DBRP as an MDP, while Section~\ref{sec:Method} presents our SPRL and DPRL methods for real-time rebalancing solutions. Section \ref{sec:experiments} provides numerical experiments and analyses, as well as an ablation study. Finally, Section~\ref{sec:Conclusion} summarizes this work and outlines future research directions.

\section{Related Work}
\label{sec:literature}

Dynamic rebalancing strategies for BSS are primarily categorized into two types: user-based rebalancing and operator-based rebalancing \citep{vallez2021challenges}. User-based rebalancing entails incentivizing users to rent or return bikes at specific stations, as discussed by \cite{haider2018inventory}. This strategy is predominantly utilized in dockless BSS. In contrast, operator-based rebalancing involves the active participation of a dedicated rebalancing fleet (commonly vehicles) to redistribute bikes, which is prevalent in station-based BSS. According to a recent statistical report by \cite{meddin2022}, station-based systems continue to be more common than dockless systems. Even when addressing rebalancing challenges for the latter, researchers tend to divide the station network into sub-clusters, each of which can be regarded as an individual station \citep[see, e.g.,][]{du2019model, luo2022dynamic, xu2018station, zhang2019dynamic}. Given the prevalence and relevance of station-based systems, our study focuses on operator-based rebalancing in station-based BSS. 

\subsection{Models based on Mixed Integer Programming} 
\label{sec3:dynamic vehicle repo}

Most existing DBRP approaches rely on MIP models with time discretization \citep[see, e.g.,][]{ghosh2019improving, mellou2019dynamic, liang2024dynamic, chiariotti2018dynamic, zheng2021repositioning}, where the temporal dimension is partitioned into fixed periods. Time discretization requires aggregating all rental and return demand for each station occurring within the same time period. This raises some issues. First, it necessitates assumptions about the sequence of rentals, returns, and rebalancing operations within each period, along with priority rules to allocate bikes when demand exceeds supply. 
Second, to ensure practical feasibility, each vehicle is typically restricted to rebalancing a maximum number of stations (typically, one) per period. 
Third, selecting an appropriate period length is non-trivial: longer periods introduce idle time and reduce accuracy, while shorter periods increase computational complexity.

In addition to such challenges, explicitly modeling the stochastic nature of trip demands would require stochastic optimization models, leading to even more complex optimization models \citep[see, e.g.,][]{lu2016robust,lowalekar2017online} and limiting the use of such models to smaller problem instances. 
While MIP models with time discretization remain widely used in BSS optimization, their limitations underscore the need for approaches that better capture the dynamic, uncertain nature of rental and return demand, as well as the asynchronous operations of the rebalancing fleet.

\subsection{Markov Decision Processes}
\label{liter:madp}

MDPs offer a natural framework for modeling DBRP as a sequential decision-making problem \citep{seo2020dynamic}, where agents, represented by vehicles, interact within a complex environment, composed of a station network with stochastic trip demands. At each step, vehicles assess the system’s state and make rebalancing decisions, typically aiming to optimize the fulfillment of user demand. Despite their relevance, MDPs have only recently gained traction in DBRP research, with a growing but still limited body of work.
A key challenge in applying MDPs to DBRP lies in the sheer size of the state space (bike availability across stations and vehicles) and action space (vehicle movements and rebalancing operations), which grow exponentially, significantly increasing computational complexity. As a result, computing an optimal policy within a reasonable time often proves impractical \citep{legros2019dynamic}.

\cite{brinkmann2019dynamic} formulated the DBRP as an MDP and introduced a dynamic lookahead policy heuristic, albeit limited to scenarios involving a single rebalancing vehicle. Building upon this work, \cite{brinkmann2020multi} extended the model to incorporate multiple vehicles, proposing a coordinated lookahead policy to simultaneously address inventory and routing decisions. \cite{legros2019dynamic} applied MDP to develop the decision-support tool for DBRP, aiming to minimize the rate of unsatisfied users who find their station empty or full. They implemented a one-step policy improvement method, which incrementally refines the strategy by focusing on the immediate next decision rather than a series of future actions, to identify priority stations. However, their approach segments the planning horizon into periods of equal length, which is subject to the same drawbacks as a discretized planning horizon typically engaged in MIP models. \cite{luo2022dynamic} designed a policy function approximation algorithm and applied the optimal computing budget allocation method to search for the optimal policy parameters for an MDP model of dynamic rebalancing with one single vehicle. 

Overall, the current literature highlights the potential of MDPs in addressing DBRP, but their application remains in early stages. Further research is needed to tackle key challenges, including computational complexity, time discretization choices, and vehicle synchronization, to improve accuracy and efficiency in decision-making.



\subsection{Reinforcement Learning}
\label{liter3:rl}

RL has made significant progress in multi-agent systems, offering promising methods for algorithmically learning effective decision-making strategies \citep{oroojlooyjadid2019review}. This is particularly relevant given the complexity of MDP encountered in DBRP, where adaptive solutions are essential. RL enables the system to learn from past experiences and dynamically adjust its strategies in response to the changing environment, shaped by the stochastic nature of user demand in BSS.

The success of RL on various optimization problems \citep{bello2016neural, kool2018attention, bengio2021machine, Liu_Fischetti_Lodi_2022, liu2022machine} motivates its application in BSS. \cite{xiao2018distributed} further highlighted that rebalancing shares many environmental characteristics (i.e., complex and high-dimensional environment, constantly changing system drivers, and closed-loop system) with gaming, finance, and marketing, where RL is typically applied.

Most RL studies in bike-sharing focus on user-based rebalancing by designing incentive mechanisms \citep{pan2019deep, duan2019optimizing, ho2021learning, schofield2021handling}. In contrast, operator-based rebalancing RL methods remain underexplored. Dynamic rebalancing requires two key decisions: the station a vehicle should visit next (routing decision), and the number of bikes to pick up or drop off upon a vehicle's arrival (inventory decision). \cite{li2018dynamic} proposed a spatio-temporal RL framework that clusters stations and trains separate RL models for each cluster to learn rebalancing policies. \cite{xiao2018distributed} proposed a distributed RL solution with transfer learning, focusing only on the inventory decision. This approach assigns a unique agent to each station, with each agent tasked with identifying and then combining the best rebalancing strategy for their respective station. This avoids an exponentially expanding action space but requires substantial computational power for parallel learning. \cite{yindeep} and \cite{yin2024deepbike} proposed an RL-based rebalancing model, using DQN with a simulator under time discretization, disregarding the sequence of rentals, returns, and rebalancing operations, resulting in a less realistic reward function. \cite{luo2022dynamic} and \cite{seo2022rebalancing} focused on the DBRP with a single vehicle, while \cite{li2018dynamic} explored multi-vehicle rebalancing in smaller station clusters (fewer than 30 stations).
 
Overall, current models either (i) focus solely on inventory decisions, overlooking the impact of routing, or (ii) attempt to make both decisions simultaneously without accounting for system state changes during inventory operations, leading to suboptimal routing choices. Simultaneous inventory and routing decisions must be carefully structured, as they can be made either upon arrival (determining inventory for the current station) or upon departure (choosing both the next station and inventory actions).

To address these limitations, we propose an RL approach with two architectures: a single-policy RL (SPRL) that jointly learns inventory and routing decisions within a single network, and a dual-policy RL (DPRL), which decouples these decisions by training two separate networks with shared state. Both models learn within a fine-grained simulator that explicitly accounts for the sequence of rentals, returns, and rebalancing operations, ensuring a realistic reward function. While SPRL provides a strong baseline---offering some improvements in scale and realism over prior work---DPRL offers a more responsive and granular solution. Specifically, upon arrival at a station, the inventory policy determines the number of bikes to be picked up or dropped off. Once the inventory operation is completed, the routing policy selects the next station to visit. This separation enhances the system’s responsiveness to dynamic changes, leading to a more realistic and effective real-time rebalancing strategy.

\section{Problem Definition and General MDP Framework}\label{sec:Model}
\subsection{Network Model}
We consider the DBRP defined over a finite planning horizon. A BSS is represented by a directed graph $G= (N,E)$ with $|N|$ stations and $E \subseteq N \times N$ arcs. A fleet $V$ of vehicles is available to rebalance bikes among the stations. All problem parameters are summarized in Table~\ref{defin of para3}. Each station $n \in N$ has a capacity $C_{n}$ of docks, and each vehicle $v \in V$ has a capacity $\hat{C}_{v}$. The initial state of the system is defined by the bike inventory $d^{n}_{0}$ at each station $n \in N$, as well as the inventory ${p}^{v}_{0}$ and location ${z}^{v}_{0}$ of each vehicle $v \in V$. Each arc $e=(i,j) \in E$ has a distance and transit time between any two stations $i \in N$ and $j \in N$, given by $D_{i,j}$ and $R_{i,j}$, respectively. Loading or unloading one bike (from the vehicle to the station, or vice-versa) is estimated to take $\beta$ minutes. The system encounters lost demand when rental or return requests cannot be satisfied due to the absence of bikes or available docks, respectively. The primary goal is to optimize rebalancing strategies for vehicles across the station network, thereby minimizing total lost demand. Parameters and variables represented in bold are tuples, matrices, or sets whose elements are tuples or matrices.

\begin{table}[!tbph]
\caption{Input parameters of the BSS rebalancing planning problem}
\label{defin of para3}
\centering
\scalebox{0.85}{
\begin{tabular}{ll}
\hline
Parameter              & Definition                                                                                        \\ \hline
$N$                    & Set of stations                                                                               \\ 
$E$                    & Set of arcs                                                                              \\ 
$V$                    & Set of vehicles                                                                               \\
$C_{n}$       & Capacity of station $n \in N$                                                                      \\
$\hat{C}_{v}$ & Capacity of vehicle $v \in V$                                                                       \\ 
$D_{i,j}$              & Distance between station $i\in N$ and $j\in N$                                                          \\
$R_{i,j}$              & Transit time between station $i\in N$ and $j\in N$                                                          \\
$\beta$              & Time (in minutes) for loading/unloading one bike                                                            \\ 
$d^{n}_{0}$   & Initial number of bikes in station $n \in N$                                      \\ 
${p}^{v}_{0}$ & Initial number of bikes in vehicle $v \in V$   \\ 
$z^{v}_{0}$   & Initial location (station) of each vehicle $v \in V$                                     \\ \hline
\end{tabular}}
\end{table}

To address this problem, we design two decision-making policies: SPRL and DPRL. In the single-policy approach, a vehicle simultaneously determines both the inventory action (i.e., the number of bikes to pick up or drop off at the current station) and its next destination upon arrival at a station. In contrast, the dual-policy approach decouples these decisions: a vehicle first selects its inventory action upon arrival and, only after completing this operation, determines its next outgoing arc. This separation allows the dual-policy framework to incorporate updated system information before making routing decisions, while the single-policy framework provides a more immediate and integrated decision-making process. The next section presents the MDP framework that applies to both approaches.

\subsection{MDP Framework for SPRL and DPRL}
\label{mdpframe}

We model the DBRP as an MDP, characterized by the tuple $(\bm{S}, \bm{A}, W, R, \gamma)$, where $\bm{S}$ is the set of system states, $\bm{A}$ is the set of actions for the agents, $W$ captures state transition probabilities, $R$ is the cumulative discounted total reward, and $\gamma$ is the discount factor, prioritizing short-term over long-term rewards. Agents operate through a sequence of steps, where each action leads to a new state and an associated reward. The key notations used in this model are summarized in Table~\ref{defin of var3}.
In this framework, vehicles act as agents that make sequential decisions to optimize rebalancing operations. Each decision step involves choosing an action, transitioning to a new state, and receiving a corresponding reward. The following sections provide a detailed breakdown of the components of the MDP model applied to DBRP.

\begin{table}[!tbph]
\caption{Notation of MMDP model for rebalancing}
\label{defin of var3}
\centering
\scalebox{0.85}{
\begin{tabular}{ll}
\hline
Symbol & Definition                                                                                                   \\ \hline
$K$                    & Sequence of steps                                                                                    \\ 
$\bm{S}$                    & Set of states                                                                                    \\ 
$T$                    & Time of each step, $T=\{t_{1},...,t_{|K|}\}$                                                                                    \\ 
$\bm{d_{k}}$   & Number of available bikes at stations, $\bm{d_{k}}=(d_{k}^{n}, \forall n \in N)$             \\
$b_{k}^{v}$ & Station where vehicle $v \in V$ is currently located or has just departed from at step $k \in K$                         \\ 
$g_{k}^{v}$ & Destination station of vehicle $v \in V$ at step $k \in K$                \\ 
$p_{k}^{v}$      & Inventory of vehicle $v \in V$ at time $t_k$  \\ 
$m_{k}^{v}$      & Estimated time until vehicle $v$ makes its next decision  \\ 
$o_{k}^{v}$      & Remaining rebalancing operations for vehicle $v \in V$ at time $t_k$ \\
$i_{k}^{v}$      & Indicator: 1 for routing state, 0 for inventory state \\
$\bm{A}$                    & Set of actions                                                                              \\ 
$l_{k}^{v}$         & Rebalancing decision for vehicle $v \in V$ at step $k \in K$\\ 
$z_{k}^{v}$         & Routing decision for vehicle $v \in V$ at step $k \in K$\\ 
$\Pi$         & Set of policies\\
$\gamma $   & Discount factor\\
\hline
\end{tabular}}
\end{table}

\subsubsection{State Space}
At each step $k\in K$, both single-policy and dual-policy formulations share a common structure of the state $\bm{S_k}=(t_k,\bm{d_k},\bm{H_k})$. 
Here, $t_{k}$ denotes the current time, $\bm{d_{k}}=(d_{k}^{n}, \forall n \in N)$ represents the inventory of each \emph{node} (station), and $\bm{H_{k}}$ encodes \emph{vehicle status on $N\cup E$}, including their locations, inventory levels, and operational status. Despite this structural similarity, the two policies differ in how vehicle states and decision-making processes are represented.

In the single policy approach, vehicle statuses are represented as $\bm{H_{k}}=(b_{k}^{v}, g_{k}^{v}, p_{k}^{v}, m_{k}^{v}, o_{k}^{v}, \forall v \in V)$,  where routing and inventory decisions are inherently coupled within a single decision step. Vehicles either arrive at a node $b^v_k\in N$ or traverse an arc $(b_{k}^{v},g_{k}^{v})\in E$ with the remaining travel time encoded in $m^v_k$. 
The inventory level of vehicle $v$ is denoted by $p_{k}^{v}$, and $o_{k}^{v}$ indicates the number of remaining rebalancing operations (i.e., bike pickups or drop-offs) to be done. The variable $m_{k}^{v}$ represents the estimated time until vehicle $v$ makes its next decision, which, in this framework, strictly corresponds to its estimated arrival time at the next station.

In contrast, the dual policy approach introduces an additional binary indicator $i_{k}^{v}$, leading to $\bm{H_{k}}=(b_{k}^{v}, g_{k}^{v}, p_{k}^{v}, m_{k}^{v}, o_{k}^{v}, i_{k}^{v}, \forall v \in V)$. The indicator $i_{k}^{v}$ differentiates between inventory and routing decisions, where $i_{k}^{v} = 0$ when vehicle $v$ arrives at a station and makes an inventory decision, and $i_{k}^{v} = 1$ once inventory operations are completed and the vehicle proceeds to a routing decision. Another key distinction lies in the interpretation of $m_{k}^{v}$: in the single-policy approach, it strictly represents the estimated arrival time at the next station since inventory and routing decisions are made simultaneously. In contrast, in the dual-policy approach, $m_{k}^{v}$ may refer either to the estimated arrival time while moving (before the next inventory decision) or the time when the current inventory operation is completed (before the next routing decision).
These differences impact the learning process and policy structure, with the dual-policy approach enabling a more modular optimization of inventory and routing tasks.

\subsubsection{Action Space}

At each step $k \in K$, an action is selected, indicating rebalancing or routing decisions. The structure of the action space differs between the single-policy and dual-policy approaches.

\noindent
\emph{SPRL action space.}
Let $\bm{A}$ denote the set of actions, where $\bm{a_{k}} \in \bm{A}$ is defined by $(l_{k}^{v}, z_{k}^{v})$. The rebalancing decision at the current station is indicated by $l_{k}^{v}$. A positive $l_{k}^{v}$ value indicates that vehicle $v$ should pick up $l_{k}^{v}$ bikes, while a negative $l_{k}^{v}$ value implies that vehicle $v$ should drop off $|l_{k}^{v}|$ bikes. The routing decision is given by $z_{k}^{v}$, indicating the next station that vehicle $v \in V$ will visit. This formulation ensures that rebalancing and routing decisions are made simultaneously upon a vehicle's arrival at a station. In our continuous-time framework, the action at step $k \in K$ is associated exclusively with a specific vehicle upon its arrival at the allocated station, allowing decisions to be made independently of the other vehicles. Feasibility enforces $(b^v_k,z^v_k)\in E$ and capacity availability:
$0\le d^{b^v_k}_k - l^v_k \le C_{b^v_k}$,\quad $0\le p^v_k + l^v_k \le \hat C_v$.

\noindent
\emph{DPRL action space.}
At each step $k \in K$, an action $\bm{a_{k}} \in \bm{A}$ corresponds either to an inventory decision or a routing decision. For inventory decisions, the action $l_{k}^{v}$ specifies the number of bikes to pick up or drop off when a vehicle arrives at a station. For routing decisions, $z_{k}^{v}$ determines the next station once a vehicle completes its inventory operation and is ready to depart. This structure enables routing and inventory decisions to be made sequentially and adapt to real-time system status.

\noindent
\emph{Predefined fill levels for inventory decisions.}
To simplify inventory decisions in both SPRL and DPRL, we rely on three predefined fill levels $\mu_{i} \in [0,1], \forall i \in \{1, 2, 3\}$, representing proportions of the station capacity. Vehicles aim to adjust station inventories to one of these levels: $\smash{\mu_{1}C_{g^{v}_{k}}}$, $\smash{\mu_{2}C_{g^{v}_{k}}}$, or $\smash{\mu_{3}C_{g^{v}_{k}}}$. The feasible inventory decisions depend on the vehicle’s capacity $\smash{\hat{C}_{v}}$, its current inventory $p^{v}_{k}$, and the station inventory $d^{g^{v}_{k}}_{k}$. The rebalancing action $l_{k}^{v}$ for vehicle $v$ at step $k$ is defined as follows:
\begin{equation}
\small
(l_{k}^{v})_{i} =
\begin{cases}
 \min \{\hat{C}_{v}-p^{v}_{k}, d^{g^{v}_{k}}_{k}-\mu_{i}C_{g^{v}_{k}}\}& \text{if}\quad \mu_{i}C_{g^{v}_{k}}< d^{g^{v}_{k}}_{k} \\ 
 \max \{-p^{v}_{k}, d^{g^{v}_{k}}_{k}-\mu_{i}C_{g^{v}_{k}}\}& \text{if}\quad  \mu_{i}C_{g^{v}_{k}}> d^{g^{v}_{k}}_{k}\\ 
 0& \text{otherwise}.
\end{cases} \label{loading}
\end{equation} 

In the first case of Equation \eqref{loading}, $l_{k}^{v} > 0$, meaning the vehicle intends to pick up $l_{k}^{v}$ bikes from the station. In the second case, $l_{k}^{v} < 0$, meaning the vehicle plans to drop off $|l_{k}^{v}|$ bikes. The third case indicates that no rebalancing operation is performed. 

\subsubsection{Network Loss (Reward) Function}

At each step $k \in K$, an action $\bm{a_{k}}$ is taken, transitioning the system to state $\bm{S_{k+1}}$. This transition generates an immediate reward $r_{k+1}$, computed based on $\bm{S_{k}}, \bm{a_{k}}$, and $ W_{k+1}$. The reward function is defined as \emph{the negative value of lost demand} across all stations during the time interval between two consecutive steps, including both lost rentals (when a user cannot find an available bike) and lost returns (when a user cannot find an available dock). This reward is computed using our event-driven simulator introduced in the next section. The objective is to maximize the cumulative expected reward, which corresponds to minimizing total lost demand over time.

The expected cumulative discounted return is given by:
\begin{equation}
R_{k}= \mathbb{E} \left[\sum_{j=0}^{K-k-1}\gamma^{j}r_{k+j+1}\right]
\end{equation}
and reflects the long-term aggregated reward from a series of actions. The discount factor $\gamma \in [0,1]$ progressively reduces the value of future rewards, prioritizing immediate rewards over distant ones. 

\subsubsection{Transition Function and Event-Driven Simulator}
\label{simulator}

At each step $k \in K$, the system transitions from state $\bm{S_{k}}$ to the next state $\bm{S_{k+1}}$ upon executing an action. This transition is governed by a stochastic transition function $W_{k+1}$, which depends on the current state $\bm{S_{k}}$, the chosen action $\textbf{a}_{k}$, and the probability of transitioning to $\bm{S_{k+1}}$. The primary source of uncertainty in this function arises from user rental and return demand, which fluctuates stochastically over time. 

To model these transitions under various demand scenarios, we develop an event-driven simulator as an interactive environment that processes individual events chronologically in continuous time, ensuring realistic system behavior. Events (rebalancing operations, rentals, and returns) are executed under the first-arrive-first-serve rule. A rental is fulfilled if the origin station has at least one available bike; otherwise, it is marked as lost demand. If the rental is fulfilled, the associated return is scheduled for its arrival time at the destination station. Similarly, a return is fulfilled if a dock is available at the destination station; otherwise, the bike is redirected to the nearest station with an available dock, and the return is recorded as lost demand. Notably, lost returns only occur if the corresponding rentals were successful. Rebalancing operations occur concurrently with customer trips and are triggered by the arrival of vehicles at stations. All events are processed in chronological order, respecting the system's temporal dynamics. 

The pseudo-code for our simulator is presented in Algorithm~\ref{algorithm3}. Our simulator processes events in continuous time, capturing the realistic interplay between user demand and dynamic rebalancing decisions.

\begin{algorithm}[!tbph]
\linespread{0.92}\selectfont
\SetAlgoLined
\SetKwInOut{Input}{Input}\SetKwInOut{Initialization}{Initialization}\SetKwInOut{Output}{Output}
\caption{Simulator with a certain action and state}\label{algorithm3}
\Input{$I$, $D_{s,s'}$, $R_{s,s'}$, $\hat{C}_{v}$, $C_{s}$, $\beta$ , $\bm{a_{k}}$ , $\bm{S_{k}}$}
\Initialization{$W$; $e_t$; $reward = 0$; 
$time = w_t(1)$; $s = w_s(1)$; $indicator = w_i(1)$; $v = w_v(1)$; $i=1$\;}
\While{$time < e_t$ and $W \neq \emptyset$}{
$sign = 0$\;
\uIf{$indicator = d$}{
\uIf{$d_{k}^{s}> 0$} 
{$d_{k}^{s} = d_{k}^{s}- 1$; $W = W \cup \{[t_a(i), s_a(i), a, 0]\}$\;} \uElse{$reward = reward - 1$\;} $i = i+1$\;} 
\uElseIf{$indicator = a$}{
\uIf{$C_{s} - d_{k}^{s} > 0$}{$d_{k}^{s} = d_{k}^{s} + 1$\;}
\uElse{$reward = reward - 1$; $s'= \argmin_{d_{k}^{s'}< C_{s'}}D_{s,s'}$ ;
    $d_{k}^{s'} = d_{k}^{s'} + 1$\;}
}
\uElseIf{$indicator = p$}{
\uIf{$d_{k}^{s} > 0$ and $p_{k}^{v} < \hat{C}_{v}$}{$d_{k}^{s} = d_{k}^{s} - 1$; $p_{k}^{v} = p_{k}^{v}+ 1$;$o_{k}^{v} = o_{k}^{v}- 1$ \;
     \uIf{$o_{k}^{v} = 0$}{$sign = 1$\;}
     }
     \uElse{Remove $w$ in $W$ whose $w_s = s, w_i = p, w_v = v$;
    $sign = 1$\;}
}
\uElse{\uIf{$p_{k}^{v} > 0$ and $d_{k}^{s} < C_{s}$}{$d_{k}^{s} = d_{k}^{s}+ 1$; $p_{k}^{v} = p_{k}^{v} - 1$; $o_{k}^{v} = o_{k}^{v}- 1$\;
     \uIf{$o_{k}^{v} = 0$}{$sign = 1$\;}
     }
     \uElse{Remove $w$ in $W$ whose $w_s = s, w_i = f, w_v = v$;
    $sign = 1$\;}}
\uIf{$sign = 1$}{$m_{k}^{v} = time + R_{s,g_{k}^{v}}$\; 
\uIf{$m_{k}^{v} < e_t$}{$e_t = m_{k}^{v}$;
Remove elements in $W$ whose $w_{i} = d$\;}}
$W = W \backslash \{[time, s, indicator, v]\}$\;
$time, s, indicator, v = w_t(m), w_s(m), w_i(m), w_v(m)$ ($m = \argmin w_t(m)$) \;}
Update $\bm{S_{k+1}}$\;
\Output{$reward$ and $\bm{S_{k+1}}$}
\end{algorithm}

As vehicle $v$ arrives at station $g_{k}^{v}$ at $t_{k}$, an action $\bm{a_{k}}$ will instantly be taken. The system status, such as station inventory and vehicle fleet information, is obtained through the current observation. A queue set $W$ is created to record individual upcoming events such as rentals, returns, and rebalancing operations. Each event in $W$ is detailed by a tuple $[w_t(m), w_s(m), w_i(m), w_v(m)]$, where $m$ is the index of each event, $w_t(m)$ specifies the time the event occurs, $w_s(m)$ identifies the involved station, $w_i(m)$ denotes the type of event ---either a rental demand (`$d$'), a return demand (`$a$'), a bike pick-up (`$p$'), or a bike drop-off (`$f$')--- and $w_v(m)$ specifies the vehicle involved, if applicable. For rental or return events, $w_v(m)$ is set to 0, as these events are not vehicle-specific.

The queue $W$ is initially populated with all rental demands and scheduled rebalancing operations (pick-ups and drop-offs) from $o_{k}^{v}$ occurring within the interval $[t_{k}, e_t)$, where $e_t$ is the estimated time of the next step derived from $m_{k}^{v}$ in Table~\ref{defin of var3}. The corresponding returns of successful rentals are created and added to $W$ in real time. All events in queue $W$ are sorted in ascending order by event time $w_t(m)$. The simulator processes the events sequentially, ensuring a first-come-first-serve consistent with real-world operations. After completing an event, the simulator removes the first element from the queue and proceeds to execute the next event accordingly.

During the simulation, the system state is continuously updated in real time, leading to the subsequent state $\bm{S_{k+1}}$ and an immediate reward. This reward is defined as the negative total lost demand incurred until the next step.


\subsection{Continuous Time Framework}
To efficiently manage the rebalancing process for both SPRL and DPRL, we adopt a continuous-time decision-making framework. This approach enables vehicles to make dynamic decisions in response to system changes, resulting in more responsive and adaptive operations.

\textbf{SPRL Time Framework.} We illustrate the SPRL time framework in Figure~\ref{fig: timeSP}. 
When vehicle $v_{i} \in V$ arrives at station $g_{k}^{v_{i}}$, all other vehicles continue executing their current actions ---either performing rebalancing operations at their respective stations or traveling to their next destinations. Upon arrival, vehicle $v_{i}$ immediately takes an action $\textbf{a}_{k}=(l_{k}^{v_{i}}, z_{k}^{v_{i}})$, determining both the inventory and routing decisions for $v_{i}$ at step $k$. Once action $\textbf{a}_{k}$ is completed, the system transitions to the next state $\bm{S_{k+1}}$, and an immediate reward is obtained within the time segment $[t_k, t_{k+1}]$ as described in Section~\ref{simulator}. The duration between decision steps is driven by system dynamics; specifically, a new time step begins whenever the next vehicle arrives at a station. 

\begin{figure}[!hbtp]
\centering
  \noindent\makebox[\textwidth]{
  \resizebox{11cm}{2.6cm}{\begin{tikzpicture}[>=Stealth,
                    node distance=2cm,
                    on grid,
                    auto,
                    every node/.style={align=center}]

   \draw[->] (0,0) -- (12,0) node[anchor=north] {};

  \node[circle,fill,inner sep=1.5pt,label=below:{$0$}] (start) at (0,0) {};
  \node[circle,fill,inner sep=1.5pt,label=below:{$k$}] (k) at (3,0) {};
  \node[circle,fill,inner sep=1.5pt,label=below:{$k+1$}] (k1) at (7.5,0) {};
  \node[circle,fill,inner sep=1.5pt,label=below:{$k+2$}] (k2) at (9,0) {};
  \node[circle,fill,inner sep=1.5pt,label=below:{$|K|$}] (end) at (11.5,0) {};

  \node[label=below:{$...$}] at (1.5,-0.1) {};
  \node[label=below:{State $\bm{S_k}$}] (sk) at (1.8,-1) {};
  \node[label=below:{Action $\bm{a_k}$}] (ak) at (4.5,-1) {}; 
  \node[label=below:{State $\bm{S_{k+1}}$}] at (7.6,-1) {};
\node[label=below:{$v_j$}] at (7.6,-0.5) {};
 \node[label=below:{$...$}] at (10.5,-0.1) {};

  \draw[->] (sk) -- (k) node[midway,below] {};
  \draw[->] (ak) -- (k) node[midway,below] {$v_i$};

\draw [thick,decorate,decoration = {brace,raise=4pt}] (3,0.1) --  (7.5,0.1) node[midway, yshift= 1em]{Immediate reward $r(\bm{S_k}, \bm{a_k})$};
\end{tikzpicture}}
  }
\caption{Continuous time planning framework of SPRL}
\label{fig: timeSP}
\end{figure}


\textbf{DPRL Time Framework.}
In previous SPRL frameworks~\citep{li2018dynamic, yindeep}, inventory and routing decisions for vehicle $v_{i}$ are made simultaneously in state $\bm{S_{k}}$. However, when vehicle $v_{i}$ completes the inventory operation at $t_{k+2}$, the routing decision made at~$t_{k}$ may no longer be optimal due to system changes, particularly user demand, that occur during the interval $[t_k, t_{k+2}]$. The DPRL framework addresses this limitation by decoupling inventory and routing decisions, capturing system dynamics in a more realistic and granular way. Specifically, DPRL executes the inventory decision when the vehicle arrives at a station and defers routing decisions until after the inventory operation is completed. This allows the routing decision to incorporate the most up-to-date system information, including recent changes in demand or station status. Similarly, routing decisions are made upon departure from the station, and inventory decisions are deferred until arrival at the next station. 

\begin{figure}[!hbtp]
\centering
  \noindent\makebox[\textwidth]{
  \resizebox{11cm}{3.2cm}{\begin{tikzpicture}[>=Stealth,
                    node distance=2cm,
                    on grid,
                    auto,
                    every node/.style={align=center}]

   \draw[->] (0,0) -- (12,0) node[anchor=north] {};

  \node[circle,fill,inner sep=1.5pt,label={[below, yshift=-0.3cm]{$0$}}] (start) at (0,0) {};
  \node[circle,fill,inner sep=1.5pt,label={[below, yshift=-0.3cm]{$k$}}] (k) at (2,0) {};
  \node[circle,fill,inner sep=1.5pt,label={[below, yshift=-0.3cm]{$k+1$}}] (k1) at (5.3,0) {};
  \node[circle,fill,inner sep=1.5pt,label={[below, yshift=-0.3cm]{$k+2$}}] (k2) at (9.3,0) {};
  \node[circle,fill,inner sep=1.5pt,label={[below, yshift=-0.3cm]{$|K|$}}] (end) at (11.5,0) {};

  \node[label=below:{$...$}] at (1,-0.2) {};
  \node[label=below:{Inventory \\State $\bm{S_k}$}] (sk) at (0.9,-1) {};
  \node[label=below:{Inventory \\Action $\bm{a_k}$}] (ak) at (3.4,-1) {}; 
  \node[label=below:{Inventory \\ State $\bm{S_{k+1}}$}] at (5.3,-1) {};
\node[label=below:{$v_j$}] at (5.3,-0.5) {};
 \node[label=below:{$...$}] at (10.7,-0.2) {};
 
 \node[label=below:{Routing \\State $\bm{S_{k+2}}$}] (sk2) at (7.8,-1) {};
  \node[label=below:{Routing \\Action $\bm{a_{k+2}}$}] (ak2) at (10.8,-1) {}; 

  \draw[->] (sk) -- (k) node[midway,below] {};
  \draw[->] (ak) -- (k) node[midway,below] {$v_i$};
  \draw[->] (sk2) -- (k2) node[midway,below] {};
  \draw[->] (ak2) -- (k2) node[midway,below] {$v_i$};

\draw [thick,decorate,decoration = {brace,raise=4pt}] (2,0.1) --  (5.3,0.1) node[midway, yshift= 1em]{Immediate reward $r^{I}(\bm{S_k}, \bm{a_k})$};
\end{tikzpicture}}
  }
\caption{Continuous time planning framework of DPRL}
\label{fig: timeDP}
\end{figure}

Figure~\ref{fig: timeDP} illustrates the state transitions in DPRL. Two distinct state types exist: inventory state and routing state, sharing the same structure described in the previous section. A decision step corresponds to the moment a vehicle either arrives at or departs from a station after completing its rebalancing operation. For instance, an inventory state $\bm{S_{k}}$ occurs when vehicle $v_{i}$ arrives at station $g_{k}^{v_{i}}$, while all the other vehicles either rebalance bikes at their stations or relocate to their next stations. At this point, an action $l_{k}^{v_{i}}$ is generated, specifying the number of bikes for $v_{i}$ to pick up or drop off at step $k$. The system then transitions to the next state $\bm{S_{k+1}}$, which may represent another inventory state for $v_{j}$ (as illustrated in Figure~\ref{fig: timeDP}) or a routing state for another vehicle. After vehicle $v_{i}$ completes the inventory rebalancing operation initiated at state $\bm{S_{k}}$, a routing decision is made in routing state $\bm{S_{k+2}}$ before it departs from the station. Here, an example of inventory reward $r^{I}(\bm{S_k}, \bm{a_k})$ is shown for the duration $[t_k, t_{k+1}]$. If the state $\bm{S_{k-1}}$ is a routing state, then the corresponding routing reward is denoted as $r^{R}(\bm{S_{k-1}}, \bm{a_{k-1}})$ for the interval $[t_{k-1}, t_{k+2}]$. 

Both SPRL and DPRL operate under a continuous-time framework, enabling vehicles to make independent decisions based on system dynamics. Nevertheless, both frameworks focus solely on the action for a single vehicle. As a result, the complexity of the action space is significantly reduced compared to the case where actions for all vehicles are taken simultaneously \citep{yindeep}. It enables each vehicle to react dynamically and instantaneously to the present state, without waiting for other vehicles. Additionally, it reflects realistic operational conditions. It is crucial to note that although there is no direct communication or coordination between the vehicles, the status of all vehicles is given within the current state, facilitating an indirect form of situational awareness. In other words, each vehicle operates autonomously, making decisions based solely on the global state of the system (i.e., knowing the status of all the other vehicles) instead of engaging in collaborative strategies with the other vehicles. 

\section{RL Policies for DBRP}
\label{sec:Method}

To optimize rebalancing decisions dynamically, we employ a Deep Q-Network (DQN)-based RL approach. This section first introduces the common RL framework underlying both the SPRL and DPRL strategies. We then detail each approach separately, highlighting key distinctions in decision-making and training methodologies.

\subsection{Q-learning and DQN}

Q-learning is a value-based, off-policy Temporal-Difference (TD) RL method. The goal is to learn a function that estimates the expected return of taking a particular action in a given state, guiding the agent toward the most rewarding outcomes. With a learned Q-value function, a greedy policy $\pi^*$ that maximizes long-term reward can be derived from Equation \eqref{oaf_ac}.
The optimal Q-function obeys the following Bellman equation \citep{bellman1957markovian}:
\begin{align}
     Q^* (\bm{S_{k}},\bm{a_{k}}) = \mathbb{E}  [r_k +\gamma \max_{\bm{a_{k+1}}} Q^*(\bm{S_{k+1}},\bm{a_{k+1}})], \label{bellman}
\end{align}
where $\bm{S_{k}}$ and $\bm{a_{k}}$ are the current state and action, $\bm{S_{k+1}}$ and $r_k$ are the next state and immediate reward after taking action $\bm{a_{k}}$, and $\bm{a_{k+1}}$ is the action that achieves maximal Q-value at state $\bm{S_{k+1}}$. The expectation is taken over the distribution of rewards and transitions.

Before the Q-value function converges to optimality, the TD error measures the discrepancy between both sides of the Bellman equation:
\begin{align}
     \delta =  r_k +\gamma \max_{\bm{a_{k+1}}}  Q(\bm{S_{k+1}},\bm{a_{k+1}})-  Q(\bm{S_{k}},\bm{a_{k}}). \label{tde}
\end{align}

Q-learning iteratively updates the Q-values by minimizing the TD error. The core update rule is given by $Q(\bm{S_{k}},\bm{a_{k}}) \leftarrow Q(\bm{S_{k}},\bm{a_{k}}) + \alpha \delta$, where $\alpha \in (0,1]$ is the learning rate. 

This update mechanism exemplifies the core idea of TD learning. It adjusts the Q-value based on the actual reward received and the estimated value of the next state. By doing so, the agent incrementally improves its predictions of future rewards. Being off-policy, Q-learning enables the agent to gain insights from exploratory actions, which may not immediately appear optimal, thereby enriching the learning process and allowing the policy to evolve beyond the limits of the agent's existing strategy.

Deep Q-Networks advance Q-learning by integrating deep neural networks as function approximators for the Q-value function. DQN employs two separate networks: a prediction network and a target network. The prediction network approximates the Q-value for any state-action pair and is updated frequently based on observed transitions. The target network, updated less frequently, provides stable target Q-values and helps prevent learning instability.

Specifically, the target Q-value for a transition is computed as $r +\gamma \max_{a'} Q(S',a';\theta^{target})$, where $S'$ is the subsequent state. The loss function used to train the prediction network is the squared TD error: 
    $L(\theta^{pred}) = \mathbb{E} \left[\left((r_k +\gamma \max_{\bm{a}} Q(\bm{S_{k+1}},\bm{a};\theta^{target})) \linebreak-Q(\bm{S_{k}},\bm{a_{k}};\theta^{pred})\right)^{2}\right]$,
where $\theta^{target}$ and $\theta^{pred}$ are the parameters of the target network and prediction network, respectively. The goal is to minimize this loss using stochastic gradient descent.

During training, the agents explore the state-action space by using an $\epsilon$-greedy strategy, where $\epsilon$ represents the exploration-exploitation trade-off, indicating the percentage of actions through which the agent takes random actions instead of following the current best policy. Initially, a high $\epsilon$ value is set, meaning the agent is more likely to take a random action, ensuring sufficient exploration. Over time, $\epsilon$ is gradually decreased, shifting the agent's behavior towards exploitation by selecting actions that maximize the Q-value according to the prediction network. As the agent interacts with the environment, its experiences ---consisting of state, action, reward, and next state--- are stored in a replay buffer. To train the network, mini-batches of these experiences are sampled randomly from the buffer, ensuring a more stable and uncorrelated training signal. The prediction network is updated at every training step, while the target network is updated less frequently, typically by copying the prediction network's weights at regular intervals. This delayed update mechanism helps stabilize training and avoids divergence or erratic behavior.

For both SPRL and DPRL, we adopt the same neural network architecture for prediction and target networks. The input layer is designed to match the dimensionality of the system state, as defined in Section~\ref{mdpframe}. This is followed by two fully connected dense layers, followed by a Rectified Linear Unit (ReLU) activation function. ReLU has been selected for its non-linear properties and its effectiveness in mitigating the vanishing gradient problem during deep network training~\citep{nair2010rectified}. Finally, the output layer contains one node per possible action, allowing the network to output a Q-value for each possible action given the current state. 

\subsection{SPRL Approach}

The pipeline of our single policy approach, i.e. SPRL is illustrated in Figure~\ref{fig: rf}. 

\begin{figure}[!hbtp]
\centering
  \noindent\makebox[\textwidth]{
  \resizebox{11.4cm}{4cm}{\begin{tikzpicture}[every node/.style={font=\Large}, 
  block/.style={
    rectangle, 
    draw, 
    fill=white,
    align=center, 
    minimum width=2.5cm, 
    minimum height=1.2cm, 
    thick
  },
  line/.style={
     -{Latex[scale=1.2]},  
    thick
  },
  background/.style={
    rectangle,
    rounded corners = 10pt,
    inner sep=0.5cm,
    draw,
    thick,
    fill=gray!20 
  }
]

\node[block] (DemandScenarios) at (0,0) {Demand\\ Scenarios};
\node[block, right=2.8cm of DemandScenarios] (Simulator) {Simulator};
\node[block, right=2.8cm of Simulator] (PolicyEvaluation) {DQN};
\node[block, right=2.8cm of PolicyEvaluation] (OnlineRebalance) {Online\\Rebalancing};

\node [label=center:\large Reward, yshift=-0.3cm, xshift=1.3cm] at (Simulator.east) (Reward) {};
\node [label=center:\large State, yshift=-1.5cm, xshift=0cm] at (Reward.south) (state1) {};
\node [label=center:\large Action, yshift=1.5cm, xshift=0cm] at (Reward.north) (action) {};

\draw[line] (DemandScenarios) -- (Simulator);
\draw[line] (Simulator) -- (PolicyEvaluation);

\draw[line] (PolicyEvaluation) -- (OnlineRebalance);
\draw[line] ($(PolicyEvaluation.north)+(-0cm,0cm)$)|- ($(Simulator.north)+(0cm,1cm)$) -- ($(Simulator.north)$) ;

\draw[line] (DemandScenarios) -- ($(DemandScenarios.north)+(0,3cm)$) -| ($(OnlineRebalance.north)+(0,0cm)$);

\draw[line] ($(Simulator.south)+(-0cm,0cm)$)|- ($(PolicyEvaluation.south)+(0cm,-1cm)$) -- ($(PolicyEvaluation.south)+(0cm,0cm)$) ;


\node[anchor=north east, yshift=2cm, xshift=1.2cm] at (PolicyEvaluation.north east) (offline){Offline Learning};

\begin{scope}[on background layer]
\node[background, fit=(Simulator)(state1)(action)(PolicyEvaluation)(offline)] (background) {};
\end{scope}

\end{tikzpicture}}
  }
\caption{SPRL Pipeline for Real-time Rebalancing}
\label{fig: rf}
\end{figure}

Building on Section~\ref{mdpframe}, we formulate our DQN model within a multi-agent framework. In SPRL, rebalancing operations are determined by the \emph{actions} taken by the agents, which depend on rental and return demand obtained from the various \emph{demand scenarios}. Given the current state, the DQN produces an action pair consisting of a rebalancing decision and an inventory decision. The simulator is responsible for computing immediate rewards between consecutive decision steps in a manner that reflects real-world conditions. Along with the current system state, these rewards are recorded and used to train our \emph{DQN}. During simulation, the system state is continually updated and fed back to the agents, forming a critical feedback loop that captures the agent-environment interaction. 

The DQN is trained through \emph{offline learning}, using a training set of demand scenarios, where user demand is associated with temporal information. While real-world demand may also be influenced by factors such as weather, our framework does not explicitly rely on such external data. Instead, it implicitly captures such variability through demand patterns in the training set. Once the offline learning phase is complete, the trained DQN encodes a rebalancing policy that is deployed during the \emph{online rebalancing} phase. In this phase, the policy is evaluated on a separate test set of demand scenarios, allowing us to assess its performance in unseen conditions.

\subsection{DPRL Approach}
\label{sec:DPRL}

DPRL pipeline (Figure~\ref{fig: rfDP}) introduces a critical architectural modification to SPRL. Unlike SPRL, DPRL explicitly decomposes the decision-making process into separate Inventory and Routing Policies. The key innovation lies in the state type segmentation, which enables more targeted learning strategies for inventory rebalancing and vehicle routing, potentially enhancing the system's adaptability and rebalancing efficiency.


\begin{figure}[!hbtp]
\centering
  \noindent\makebox[\textwidth]{
  \resizebox{12.5cm}{6cm}{\begin{tikzpicture}[every node/.style={font=\Large}, 
  block/.style={
    rectangle, 
    draw, 
    fill=white,
    align=center, 
    minimum width=2.5cm, 
    minimum height=1.2cm, 
    thick
  },
  decision/.style={
    diamond,
    draw,
    fill=white,
    align=center,
    aspect=2,
    minimum width=2.5cm,
    minimum height=1.2cm,
    thick
  },
  line/.style={
     -{Latex[scale=1.2]},  
    thick
  },
  background/.style={
    rectangle,
    rounded corners = 10pt,
    inner sep=0.4cm,
    draw,
    thick,
    fill=gray!20 
  }
]

\node[block] (DemandScenarios) at (0,0) {Demand\\ Scenarios};
\node[block, right=2.1cm of DemandScenarios] (Simulator) {Simulator};
\node[decision, right=0cm of Simulator,  yshift=-3cm] (Decision) {State \\ Type};
\node[block, right=4.2cm of Simulator, yshift=0.4cm] (InventoryPolicy) {Inventory\\ Policy};
\node[block, below=0.1cm of InventoryPolicy] (RoutingPolicy) {Routing\\Policy};
\node[anchor=north, yshift=0.8cm, xshift=0cm] at (InventoryPolicy.north)(Dual) {Dual Policy};
\node[block, draw, fit={(InventoryPolicy) (RoutingPolicy)(Dual)}, inner sep=0.2cm, fill opacity=0] (PolicyEvaluation) {};

\node[block, right=9cm of Simulator] (OnlineRebalance) {Online\\Rebalancing};

\node [label=center:\large Reward, yshift=0.3cm, xshift=1.8cm] at (Simulator.east) (RewardI) {};
\node [label=center:\large State, yshift=-1cm, xshift=-0.5cm] at (Simulator.south) (stateT) {};
\node [label=center:\large Action, yshift=1.8cm, xshift=0cm] at (RewardI.north) (action) {};

\draw[line] (DemandScenarios) -- (Simulator);
\draw[line] ($(Simulator.south)+(-0cm,-0cm)$) |- ($(Decision.west)+(-0cm,-0.0cm)$);
\draw[line] ($(Simulator.east)+(-0cm,0cm)$) -- ($(PolicyEvaluation.west)$);
\draw[line] ($(Decision.east)+(-0cm,-0cm)$) -| ($(PolicyEvaluation.south)+(-0cm,-0.0cm)$);
\draw[line] (PolicyEvaluation) -- (OnlineRebalance);

\draw[line] ($(PolicyEvaluation.north)+(-0cm,0cm)$)|- ($(Simulator.north)+(0cm,2cm)$) -- ($(Simulator.north)$) ;

\draw[line] (DemandScenarios) -- ($(DemandScenarios.north)+(0,4cm)$) -| ($(OnlineRebalance.north)+(0,0cm)$);



\node[anchor=north east, yshift=2.8cm, xshift=0.9cm] at (InventoryPolicy.north east) (offline){Offline Learning};

\begin{scope}[on background layer]
\node[background, fit=(Simulator)(action)(RewardI)(Decision)(InventoryPolicy)(offline)] (background) {};
\end{scope}

\end{tikzpicture}}
  }
\caption{DPRL Pipeline for Real-time Rebalancing}
\label{fig: rfDP}
\end{figure}

\textbf{Dual DQN.}
To derive an optimal policy for dynamic rebalancing, we employ two DQNs ---one for inventory decisions and one for routing decisions. Given the structural similarities between these tasks, both networks share the same neural architecture, differing only in their output layers. The input layer aligns with the dimensions of state observations defined in Section~\ref{mdpframe}. Subsequently, two fully connected dense layers are followed by Rectified Linear Unit (ReLU) activation functions. The output layer is tailored to the action space of each network, allowing the network to generate Q-values for all possible actions based on a given state. Specifically, the action space of the inventory network comprises three fill levels, while the action space of the routing network encompasses the number of stations.

\textbf{Heuristic Initialization.}
To accelerate learning and guide early-stage training, we introduce a heuristic routing policy as the initial strategy for the routing DQN. This heuristic balances spatial proximity and inventory imbalance, assigning relatively full stations to relatively empty vehicles (and vice versa), while considering travel distance between stations. We define a routing distribution $u(n)$, representing the probability of selecting station $n$ when vehicle $i$ departs from its current station $x$ at step $k$:
\begin{align}
\small
\label{routing_heur}
u(n) = \sigma \rho_{1}(n) + (1-\sigma) \rho_{2}(n),
\end{align}
\begin{align}
\text{where} \quad \rho_{1}(n) &= (1/D_{x, n})^m/\sum_{n'}  (1/D_{x, n'})^m \label{distance} \\
\text{and} \quad \rho_{2}(n) &= [g(n)]^m/\sum_{n'} [g(n')]^m, \\ 
& g(n) = \frac{C_{n}-d_{k}^{n}}{C_{n}} \cdot \frac{p_{k}^{i}}{\hat{C}_{i}} + \frac{d_{k}^{n}}{C_{n}} \cdot \frac{\hat{C}_{i}- p_{k}^{i}}{\hat{C}_{i}}.  \label{capacity}
\end{align}
In these equations, $\rho_{1}(n)$ represents the normalized and exponentiated inverse distance between station $n$ and the departing station $x$. This ensures higher probabilities for nearby stations, reflecting the intuitive preference for shorter travel distances. The term $\rho_{2}(n)$ captures the inventory compatibility between station $n$ and vehicle $i$. As defined in Equation~(\ref{capacity}), the function $g(n)$ estimates the suitability of station~$n$ as a drop-off or pick-up location based on its current inventory level and the vehicle's available capacity. If vehicle $i$ is relatively full, stations with lower inventory (i.e., in need of bikes) receive higher $\rho_{2}(n)$ values, increasing their likelihood of being selected, and vice versa. 

The weight $\sigma \in [0,1]$ adjusts the influence of these two factors, enabling a flexible focus on different aspects of the rebalancing problem. The exponent $m$ controls the distribution sharpness: as
$m \rightarrow 0$, routing becomes uniformly randomized; as $m= \infty$, the policy becomes greedy, always selecting the highest scoring station according to these metrics.

\section{Computational Experiments}
\label{sec:experiments}
In this section, we assess the performance of our proposed RL-based approaches in comparison to several MIP and RL baselines. We first describe the datasets used for training and evaluation, followed by the benchmark methods, implementation details, and experimental results.


\subsection{Dataset}
\label{sec:dataset}

Although real-world trip data is available, we opt to use synthetic instances imitating real conditions due to several considerations. Real-world data lacks information on unobserved demand, contains inconsistencies or noise, and omits operator rebalancing operations, making station inventories unreliable. To address these issues, we use synthetic trip data instances based on varying weather conditions and temporal features, allowing RL models to learn a rebalancing policy for diverse demand scenarios. We consider two ground-truth datasets, GT1 and GT2, each with a 60-station network and varying station distributions. In GT1, nine stations are located within one city center—a high-density urban zone with larger stations that typically attract inbound commuting traffic during morning peak hours—whereas GT2 features twelve stations across two city centers. City center stations have $40$ docks, while those outside have $20$ docks. The inclusion of two configurations allows us to assess rebalancing performance under different spatial demand patterns. GT2 reflects more balanced commuting patterns, with a higher proportion of work-related trips distributed between suburban and urban stations, leading to lower demand stress overall. The inclusion of both datasets is essential for validating algorithm robustness under varying urban mobility conditions. For each dataset, there are 150 days of trip data (origin station, departure time, destination station, and arrival time). The first 100 days in each dataset are used for training, while the remaining 50 days form the test set. A fleet of four rebalancing vehicles, each with a capacity of 40 bikes, is available for repositioning bikes across stations. The details of the datasets can be found in \cite{liang2024dynamic2}.

\subsection{Benchmarks}
\label{sec:baselines}

We evaluate SPRL and DPRL against various MIP and RL models. 
\begin{itemize}

\item \textbf{Static Rebalancing (SR):} This model uses the MIP approach from \cite{liang2024dynamic} to optimize initial station inventories at the start of each day. During simulation, user demand unfolds without any further rebalancing, making SR a baseline static solution. Importantly, this model also provides the initial inventory configuration used by all other approaches.

\item \textbf{Dynamic Rebalancing (DR):} This method applies the multi-period dynamic rebalancing MIP model from \cite{liang2024dynamic}, using time discretization with 30-minute (\textbf{DR30}) and 60-minute (\textbf{DR60}) intervals. It generates rebalancing strategies indicating how many bikes each vehicle should pick up or drop off at specific stations within each time period, based on average rental and return patterns observed in the training set.

\item \textbf{RL Inventory with Heuristic Routing (RIHR):}
In this baseline, an RL model is trained only for inventory decisions upon a vehicle's arrival at a station. The routing decision is determined using a fixed greedy heuristic rule with $m=\infty$. Notably, this heuristic routing rule is also used in DPRL but only for initialization (see Section~\ref{sec:DPRL}). Unlike RIHR, DPRL gradually learns routing decisions, moving beyond the initial heuristic.


\item \textbf{RL Routing with Heuristic Inventory (RRHI):} In this baseline, an RL model is trained only for routing decisions, learning which station to visit next. The inventory decision is determined by a target inventory level, which is computed based on a demand prediction model from \cite{liang2024dynamic2}. When a vehicle arrives at a station, it aims to rebalance the inventory to this predefined target level.

    
\end{itemize}

RIHR and RRHI both partially rely on heuristic rules, while SPRL and DPRL jointly learn both inventory and routing decisions, enabling more adaptive and efficient rebalancing strategies.

\subsection{Experimental Results}
\label{sec:results}

We evaluate all models over a 4-hour planning horizon, from 7:00 a.m. to 11:00 a.m., covering the morning peak demand period.  To assess the practical efficacy and robustness of our approaches, we compare their performance against the benchmark models described in Section~\ref{sec:baselines} on the test set.

Figure~\ref{fig:mainresults} reports the average episodic lost demand, along with its standard deviation, on the test set, for each method on both GT1 and GT2 datasets. For RL-based models, we evaluate two settings based on the value of the exploration parameter $\epsilon$, which denotes the probability of selecting a random action. When $\epsilon = 0$, actions are chosen deterministically based on the highest Q-value from the trained network. When $\epsilon = 0.05$, a stochastic component is introduced, allowing a 5\% chance of taking a random action. This helps assess the robustness of the learned policies by enabling occasional exploration during evaluation. Such randomness can reveal more effective strategies that might have been missed during training and help prevent overfitting by occasionally deviating from the deterministic policy. Since MIP models are deterministic and do not rely on exploration parameters, their performance is only reported for $\epsilon = 0$. 

\begin{figure}[!hbtp]
    \centering
    \vspace{-0.5\baselineskip}
      \noindent\makebox[\textwidth]{
      \resizebox{7cm}{6.7cm}{\definecolor{purple}{HTML}{8535E5}
\definecolor{yellow}{HTML}{EDB448}
\pgfplotsset{
/pgfplots/my legend/.style={
legend image code/.code={
\draw[thick,black](-0.05cm,0cm) -- (1cm,0cm);%
   }
  }
}
\begin{tikzpicture}
    \begin{axis}[
        x=0.7cm,
        width  = \textwidth,
        height = 7cm,
        title = \tiny (A) GT1,
        ybar=2*\pgflinewidth,
        bar width=7pt,
        ymajorgrids = true,
        ylabel = {\scriptsize Total average lost demand},
        symbolic x coords={A,B,C,D,E,F,G},
        xtick ={A,B,C,D,E,F,G},
        xticklabels={SR, DR60, DR30, RIHR, RRHI, SPRL, DPRL},
        scaled y ticks = false,
        enlarge x limits= 0.1, 
        ymin=5, ymax=140,
        legend pos=north east,
        nodes near coords,
        every node near coord/.append style={font=\tiny, yshift=2pt, 
        /pgf/number format/.cd,
            fixed,
            fixed zerofill,
            precision=1},
        tick label style={font=\tiny},
        ylabel near ticks, 
        ylabel shift=-6pt ]

        \addplot[
            style={black,fill=white,mark=none},
            error bars/.cd,
            y dir=both, y explicit,
            error bar style={color=black}
        ] coordinates {
            (A,112.8) +- (0,26.7)
            (B,56.8)  +- (0,14.6)
            (C,44.8)  +- (0,15)
            (D,52.1)  +- (0,10)
            (E,59.4)  +- (0,17)
            (F,35.12)  +- (0,11.46)
            (G,23.08) +- (0,5)
        };
        
        \addplot[
            style={black,fill=gray,mark=none},
            error bars/.cd,
            y dir=both, y explicit,
            error bar style={color=black}
        ] coordinates {
            (A,0)     +- (0,0)
            (B,0)     +- (0,0)
            (C,0)     +- (0,0)
            (D,53.9)  +- (0,10)
            (E,60.0)  +- (0,17)
            (F,42.7)  +- (0,15.64)
            (G,25.89) +- (0,9)
        };

        \legend{\scriptsize $\epsilon = 0.00$, \scriptsize $\epsilon = 0.05$}
        
    \end{axis}
\end{tikzpicture}}
      \resizebox{7cm}{6.7cm}{\definecolor{purple}{HTML}{8535E5}
\definecolor{yellow}{HTML}{EDB448}

\pgfplotsset{
/pgfplots/my legend/.style={
legend image code/.code={
\draw[thick,black](-0.05cm,0cm) -- (1cm,0cm);%
   }
  }
}
\begin{tikzpicture}
\begin{axis}[
        x=0.7cm,
        width  = \textwidth,
        height = 7cm,
        title = \tiny (B) GT2,
        ybar=2*\pgflinewidth,
        bar width=6pt,
        ymajorgrids = true,
        ylabel = {\scriptsize Total average lost demand},
        symbolic x coords={A,B,C,D,E,F,G},
        xtick ={A,B,C,D,E,F,G},
        xticklabels={SR, DR60, DR30, RIHR, RRHI, SPRL, DPRL},
        scaled y ticks = false,
        enlarge x limits= 0.1, 
        ymin=5, ymax=140,
        legend pos=north east,
        nodes near coords,
        every node near coord/.append style={font=\tiny, yshift=2pt, 
        /pgf/number format/.cd,
            fixed,
            fixed zerofill,
            precision=1},
        tick label style={font=\tiny},
        ylabel near ticks, 
        ylabel shift=-6pt ]

        \addplot[
            style={black,fill=white,mark=none},
            error bars/.cd,
            y dir=both, y explicit,
            error bar style={color=black}
        ] coordinates {
            (A,82.12)  +- (0,21)
            (B,75.32)  +- (0,17)
            (C,36.72)  +- (0,15)
            (D,22.18)  +- (0,6)
            (E,32.73)  +- (0,13.6)
            (F,29.54)  +- (0,7.5)
            (G,9.99)   +- (0,4.8)
        };
        
        \addplot[
            style={black,fill=gray,mark=none},
            error bars/.cd,
            y dir=both, y explicit,
            error bar style={color=black}
        ] coordinates {
            (A,0)     +- (0,0)
            (B,0)     +- (0,0)
            (C,0)     +- (0,0)
            (D,26.07) +- (0,8.6)
            (E,34.74) +- (0,14.8)
            (F,31.05) +- (0,10.5)
            (G,12.295)+- (0,5.25)
        };

        \legend{\scriptsize $\epsilon = 0.00$, \scriptsize $\epsilon = 0.05$}
        
    \end{axis}
\end{tikzpicture}}
      }
    \vspace{-\baselineskip}
    \caption{Total average lost demand on the test set for GT1 and GT2 }
    \label{fig:mainresults}
\end{figure}

As observed in these experiments, the SR model consistently incurs the highest lost demand, with 112.82 for GT1 and 82.1 for GT2, highlighting its inability to adapt to dynamic changes throughout the day. In contrast, dynamic rebalancing strategies, such as DR30 and DR60, improve significantly over SR by incorporating user demand information throughout the day. However, they still underperform compared to several RL models.

Among the RL models, SPRL performs competitively relative to RIHR and RRHI, as it integrates both inventory and routing decisions.
Nonetheless, SPRL is limited by its concurrent decision-making approach, which does not account for system changes during vehicle inventory operations. DPRL demonstrates superior adaptability and effectiveness in minimizing lost demand, achieving the lowest lost demand values of 23.1 for GT1 and 10.0 for GT2. Specifically, DPRL reduces lost demand by 48.4\% for GT1 and 72.8\% for GT2 compared to DR30, and by 34.2\% for GT1 and 66.1\% for GT2 compared to SPRL. By decoupling inventory and routing decisions, DPRL highlights the effectiveness of a more realistic and granular framework for DBRP. 

\begin{figure}[!htbp]
    \centering
    \begin{minipage}{0.48\linewidth}
        \centering
        \includegraphics[width=\linewidth]{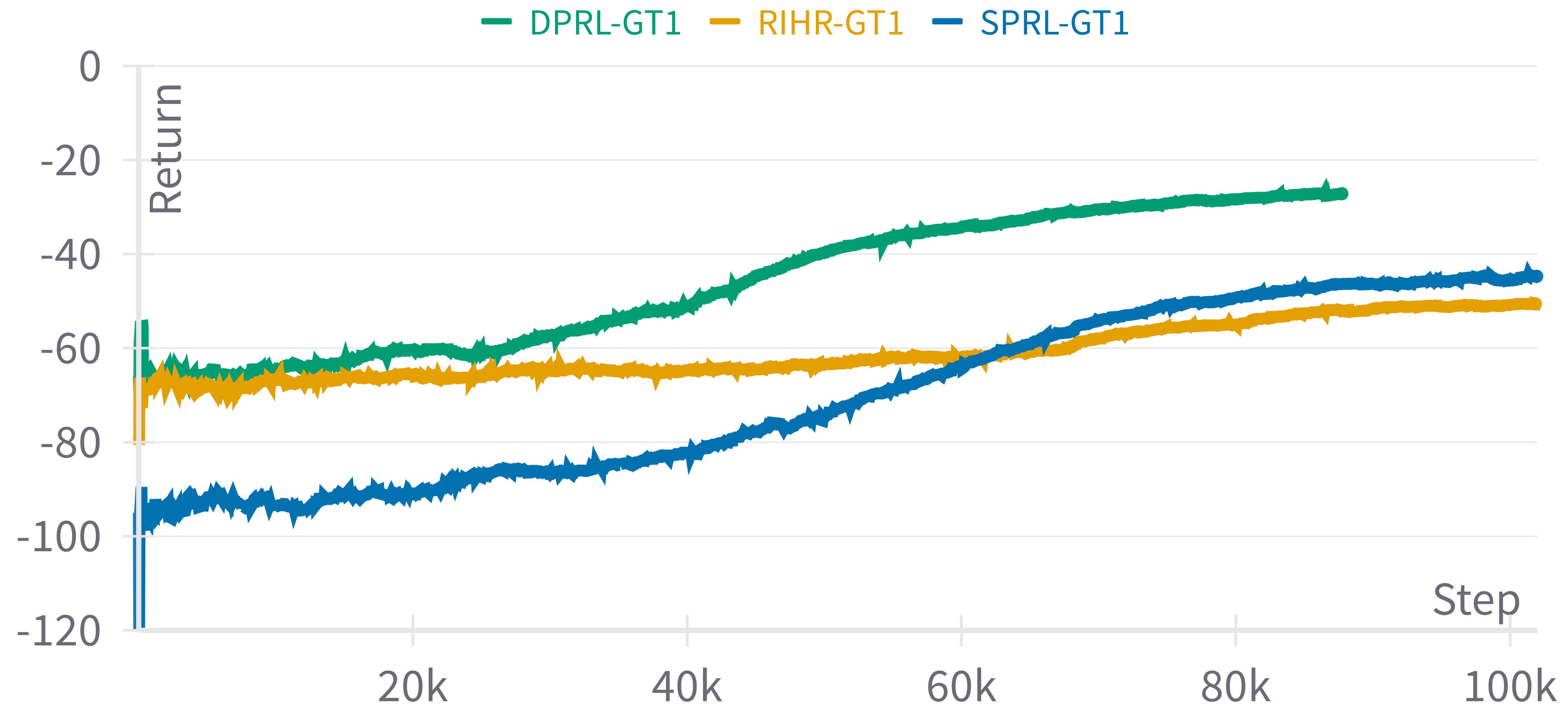}
    \end{minipage}
    \hfill
    \begin{minipage}{0.48\linewidth}
        \centering
        \includegraphics[width=\linewidth]{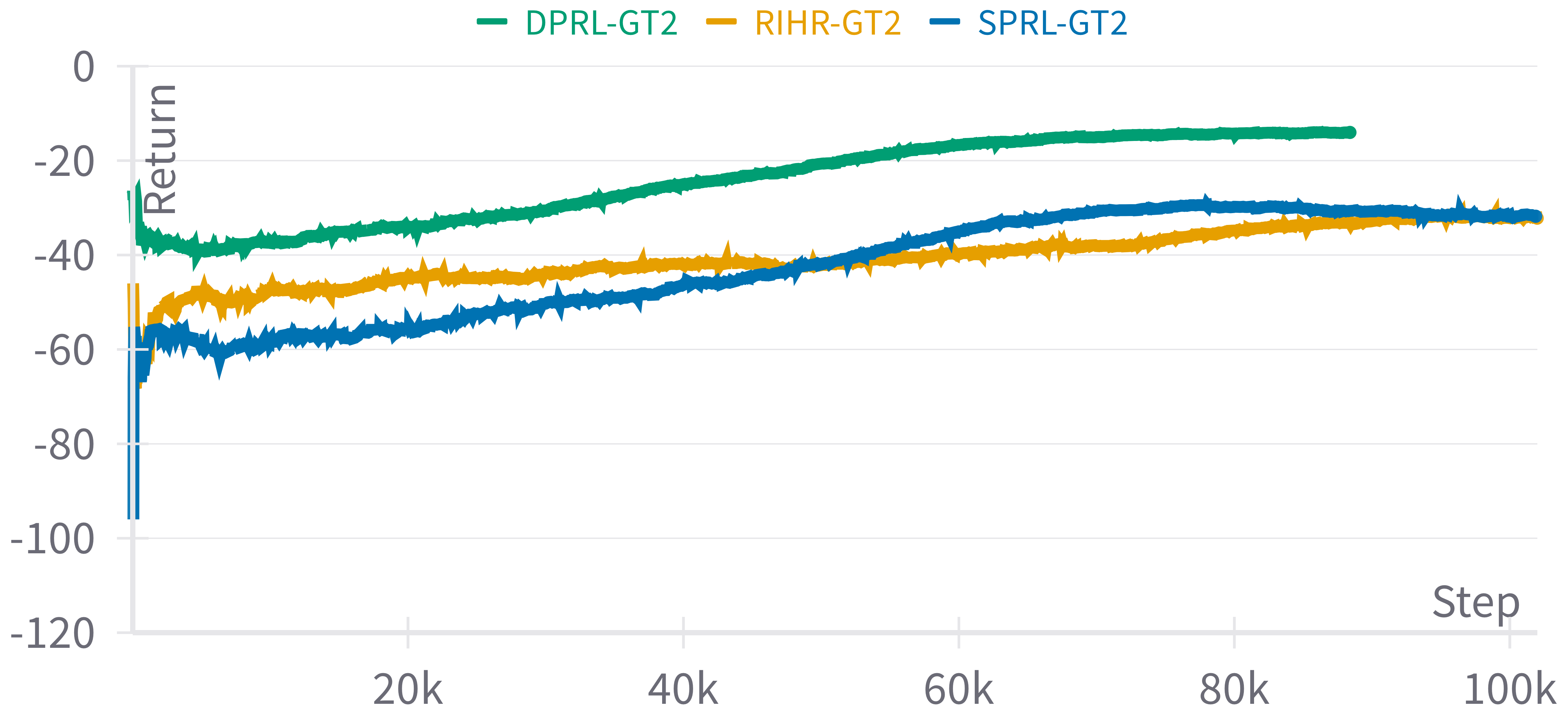}
    \end{minipage}
    \vspace*{0.2cm}
    \caption{Episodic return during RL training for GT1 and GT2}
    \label{fig:return}
\end{figure}

To comprehensively understand the learning process, Figure~\ref{fig:return} presents the episodic return during training for the three top-performing RL models. Each step corresponds to one training episode, which spans the full planning horizon of four hours. The episodic return measures the cumulative reward obtained within an episode, with higher values indicating more successful rebalancing strategies.

As shown in Figure~\ref{fig:return}, all models exhibit a general upward trend in episodic return over the course of training, indicating that they are learning progressively better rebalancing strategies. Among them, the DPRL model shows a notably steeper and more stable increase, suggesting faster convergence and more consistent improvement.

\begin{figure}[!htbp]
    \centering
    \begin{minipage}{0.48\linewidth}
        \centering
        \includegraphics[width=\linewidth]{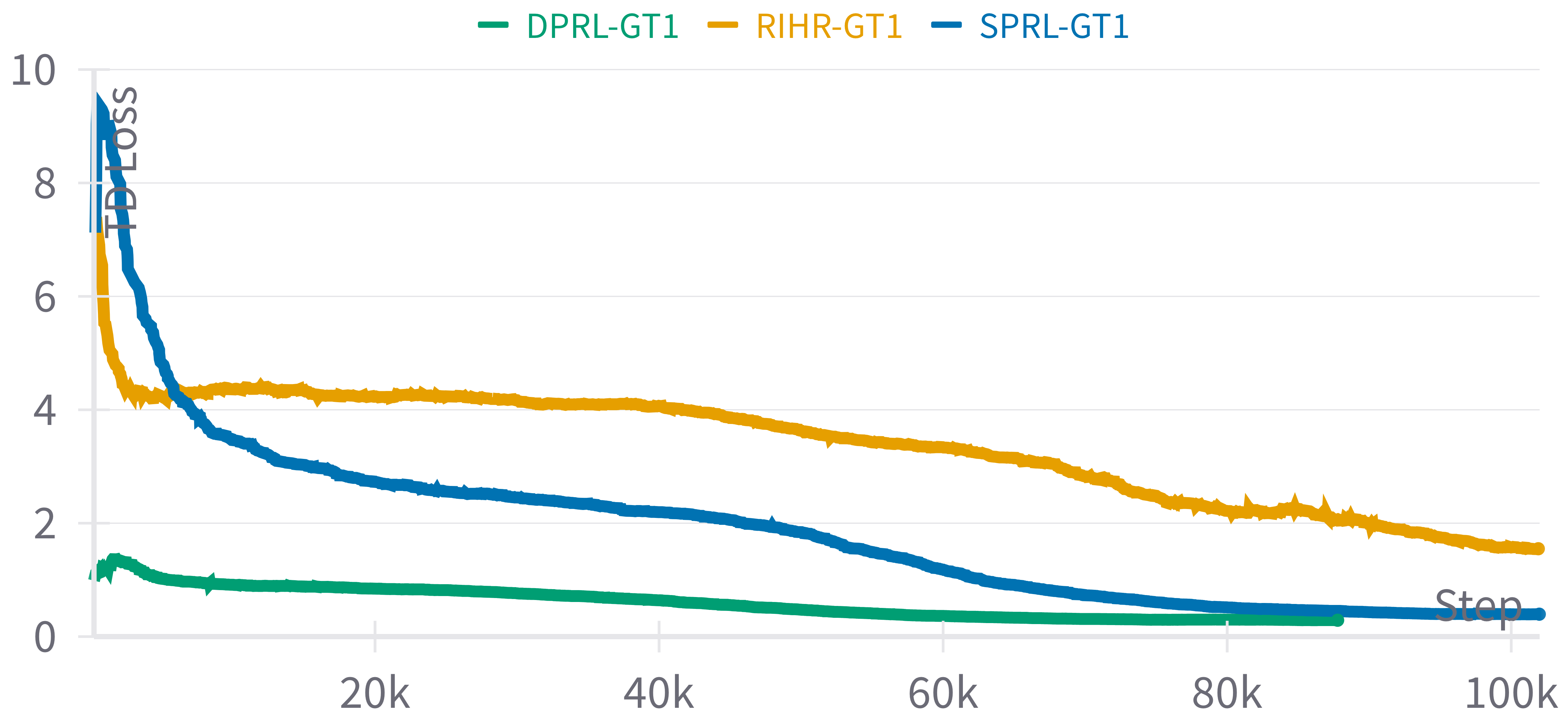}
    \end{minipage}
    \hfill
    \begin{minipage}{0.48\linewidth}
        \centering
        \includegraphics[width=\linewidth]{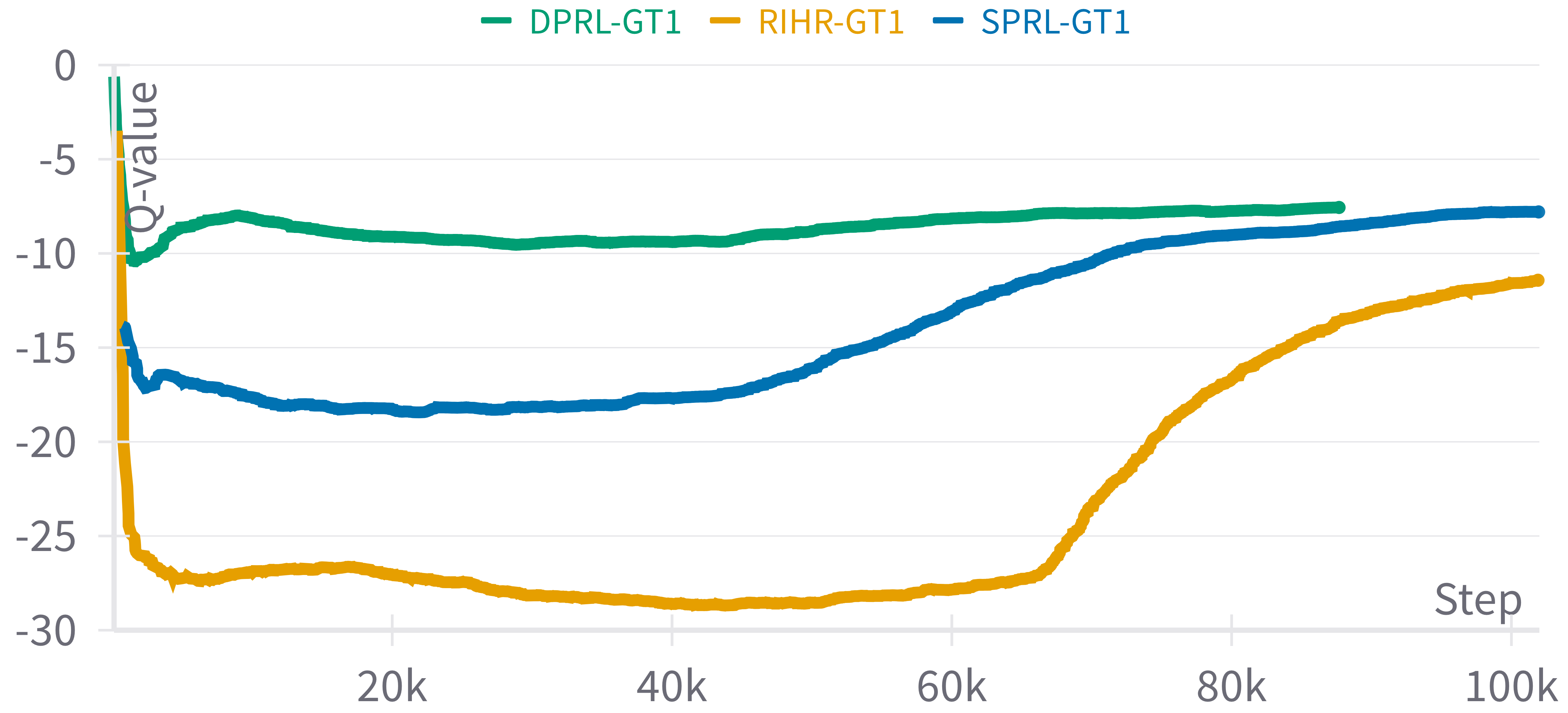}
    \end{minipage}
        \vspace*{0.2cm}
    \caption{TD loss and Q-value during RL training for GT1}
    \label{fig:GT1Train}
\end{figure}

\begin{figure}[!htbp]
    \centering
    \begin{minipage}{0.48\linewidth}
        \centering
        \includegraphics[width=\linewidth]{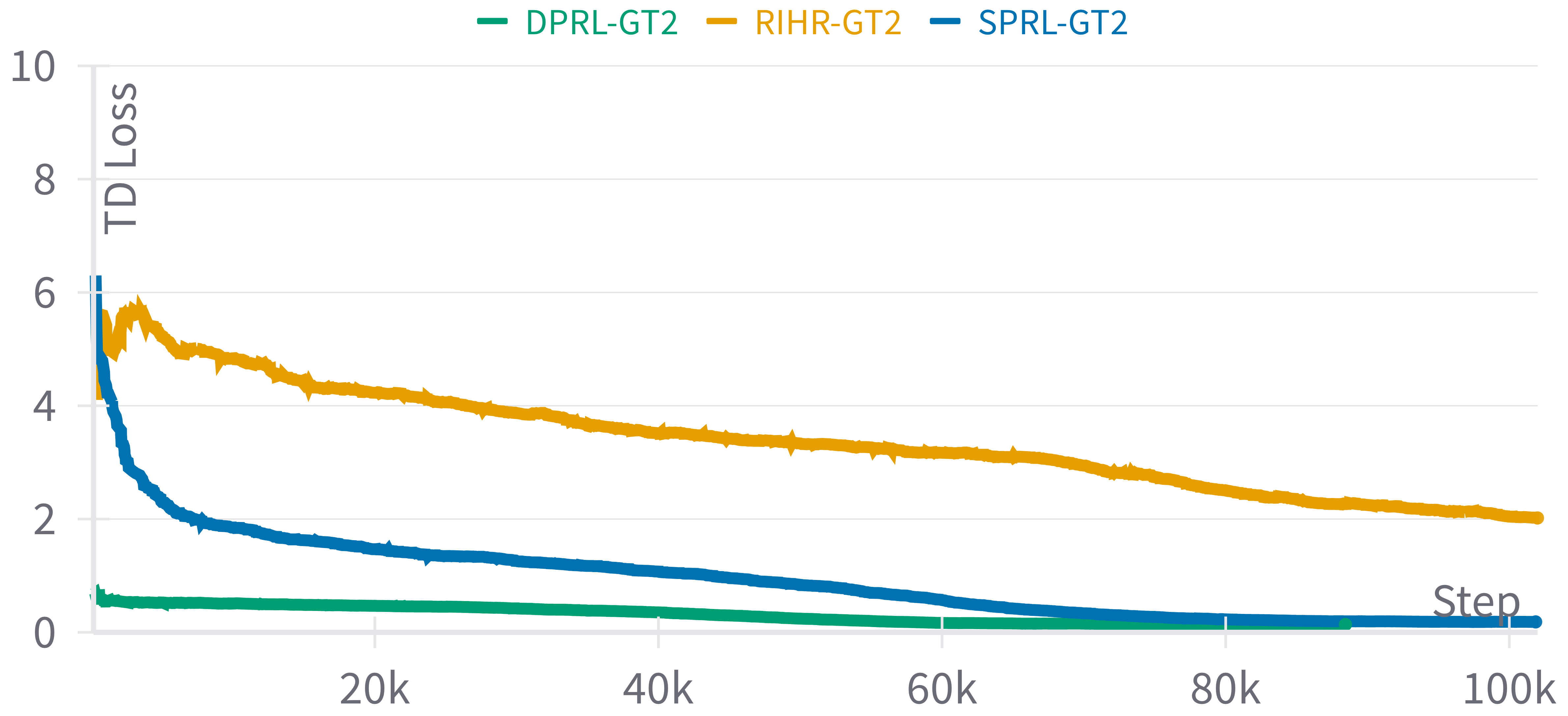}
    \end{minipage}
    \hfill
    \begin{minipage}{0.48\linewidth}
        \centering
        \includegraphics[width=\linewidth]{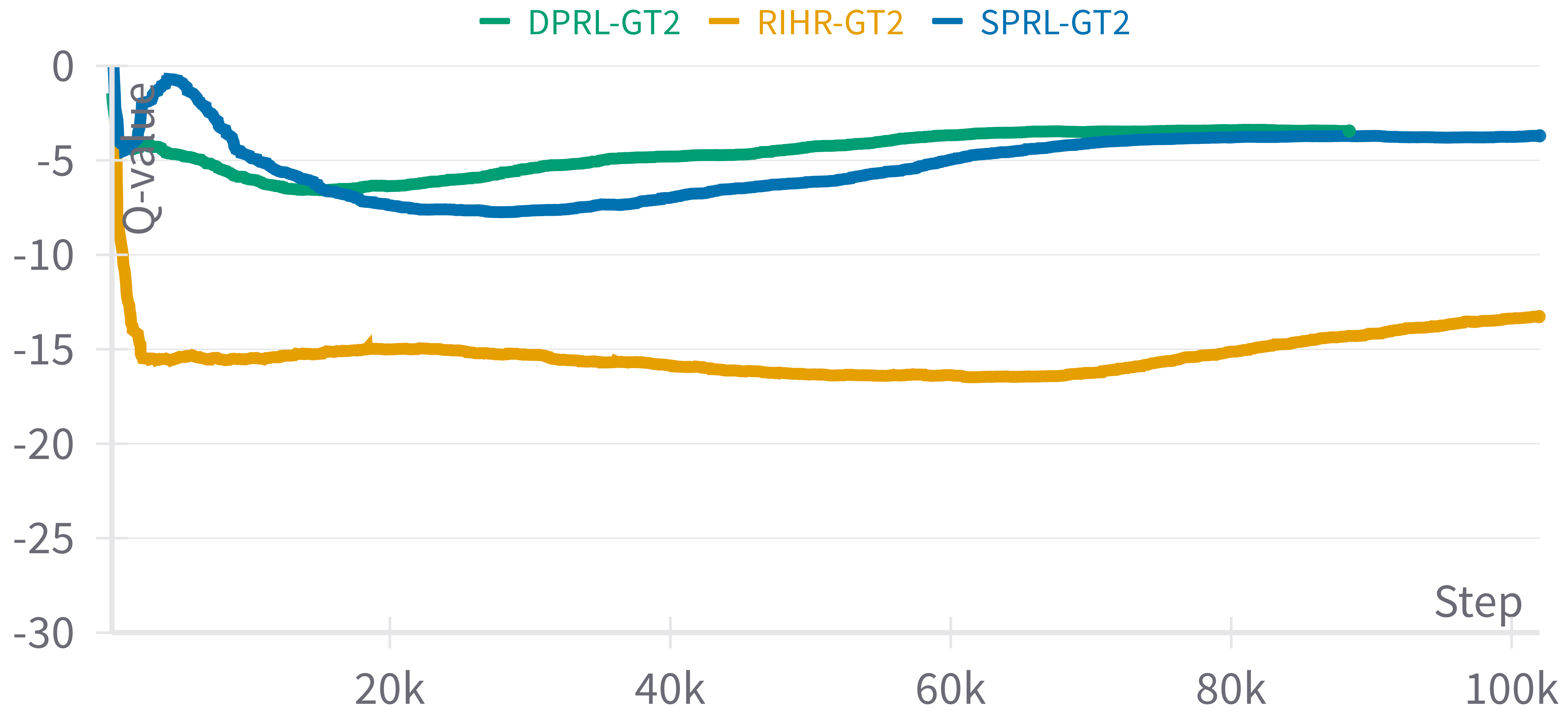}
    \end{minipage}
        \vspace*{0.2cm}
    \caption{TD Loss and Q-value during RL training for GT2}
    \label{fig:GT2train}
\end{figure}

\textbf{TD Loss and Q-value.} To further assess the training dynamics of our approach, we report the TD loss and the Q-value evolution of the three models for both GT1 and GT2, as shown in Figure~\ref{fig:GT1Train} and Figure~\ref{fig:GT2train}. The TD Loss quantifies the divergence between the predicted and the target Q-values, serving as an indicator of the network’s precision in forecasting future rewards. The Q-value reflects the expected cumulative reward for selecting a given action in a state and following the learned policy thereafter. An increase in Q-value over time signifies that, as training progresses, higher quality actions are selected, reducing expected lost demand. 

As illustrated in the figures, the TD Loss consistently decreases during training for all RL models, confirming overall learning improvement. Notably, DPRL maintains a lower TD Loss throughout, suggesting faster value-function learning. Initially, the Q-value of all RL models decreases as agents explore the state-action space with a random policy. As agents learn from these experiences, Q-values increase steadily, leading to better rebalancing policies. DPRL ultimately converges to a higher Q-value than the baselines, demonstrating its ability to guide the agent toward behaviors that maximize cumulative rewards.

Training each RL model takes approximately 18 hours on a single GPU, which is acceptable given that training is conducted offline. Once trained, the inference time is negligible, enabling real-time decision-making during operations. Regarding scalability, larger systems could be handled by clustering stations and training policies within each cluster, a common strategy in the literature that balances performance and computational tractability. From an operational standpoint, DPRL's 72.8\% reduction in lost demand represents a very substantial performance gain. In high-demand scenarios similar to our GT2 dataset, such improvements could meaningfully reduce revenue loss from unmet customer demand and likewise help increase user satisfaction. The approach's ability to learn from historical patterns while adapting to real-time conditions suggests promising applicability for practical deployment.

\section{Conclusions}
\label{sec:Conclusion}

In this study, we proposed two RL frameworks, Single-Policy RL (SPRL) and Dual-Policy RL (DPRL), to address the Dynamic rebalancing problem in BSS network. Both models operate in a continuous-time framework, allowing multiple vehicles to make independent and cooperative decisions while dynamically adapting to real-world system changes. Our overarching goal was to develop efficient rebalancing strategies that minimize lost demand and improve BSS operations beyond traditional optimization-based approaches.

SPRL formulates the DBRP as a Markov Decision Process and directly determines inventory and routing decisions upon vehicle arrival. Using a DQN trained offline, SPRL enables instantaneous rebalancing decisions in real time. Our results demonstrate that SPRL significantly outperforms MIP models, achieving up to  21.7\% reduction in lost demand compared to the best-performing MIP. Building on SPRL, the DPRL framework decouples inventory and routing decisions, enhancing decision granularity and enabling more flexible and realistic rebalancing strategies. DPRL consistently achieves superior performance across diverse scenarios, further reducing lost demand by up to 66.1\% compared to SPRL and 72.8\% compared to the best-performing MIP model. By modeling rebalancing through distinct inventory and routing subproblems, DPRL captures network dynamics more accurately and ensures vehicles make informed decisions without waiting for global updates.

Together, SPRL and DPRL illustrate the strong potential of RL-based approaches for dynamic rebalancing in BSS. They offer a scalable, computationally efficient approach that shows strong potential for improving real-world bike-sharing operations, with the dual-policy framework providing a foundation for future practical implementations. The research perspectives connected to our work are numerous. One direction involves scaling up our models to larger BSS networks and considering the specificities of E-bike sharing systems (due to charging decisions). Additional model extensions could focus on including real-time traffic data, context-aware user behavior models, or dynamic pricing incentives. Lastly, the methodological framework introduced here can be applied more broadly to other balancing and dispatch problems in emerging mobility-on-demand networks, including autonomous vehicle fleets,  \citep[see, e.g.][]{Jungel2023}.

\bibliographystyle{IEEEtranSN}
\bibliography{sample}

\newpage
\appendix 

\section{Experiment Settings and Hyperparameters}
All the RL models mentioned above are evaluated on the test set after training. The MIP models are solved using IBM ILOG CPLEX v20.1.0.0 on 2.70 GHz Intel Xeon Gold 6258R machines with 8 cores, terminating when the optimization gap falls below 0.01\% or after 24 hours. The RL models are trained using a single Tesla V100-PCIE-32GB GPU. The hyperparameters used in our RL algorithms are summarized in Table~\ref{tab:para}.

\begin{table}[!tbhp]
\centering
\caption{Parameters in training process}
\label{tab:para}
\scalebox{0.85}{
\begin{tabular}{l|r}
\hline
Parameters                  & Values    \\ \hline
Total time step             & 3,000,000   \\
Learning rate               & 2.5e-4    \\
Buffer size                 & 10,000     \\
Discount factor $\gamma$                    & 0.99      \\
Batch size                  & 256       \\
Exploration rate $\epsilon$ & 1 $\rightarrow$ 0.05 \\
Exploration fraction        & 0.5       \\
1st layer neuron            & 1,024      \\
2nd layer neuron            & 512      \\ \hline
\end{tabular}}
\end{table}

\section{Ablation Analysis}
\label{sec:ablation}

\textbf{Training Initialization.} In this section, we conduct an ablation study to examine the impact of different training initializations for DPRL. The analysis focuses on the GT1 dataset and examines how varying the parameter $m$ in Equations~\eqref{distance}~and~\eqref{capacity} ---used to define the heuristic routing distribution during initialization--- affects final model performance.
 Table~\ref{lostdemand} reports the average lost demand (mean $\pm$ standard deviation) on the test set for different values of $m$. Here, $m = \infty$ corresponds to a greedy heuristic routing rule, where the fullest vehicles are assigned to the emptiest stations, and vice versa.

\begin{table}[!tbhp]
\centering
\caption{Total average lost demand (mean $\pm$ standard deviation) on test set with varying $m$}
\label{lostdemand}
\scalebox{0.85}{
\begin{tabular}{l|ll}
\hline
Initialization with $m$ & $\epsilon = 0.00$ & $\epsilon = 0.05$ \\ \hline
$m=0$                   & $19.66   \pm 6.51$       & $28.03  \pm 14.84$           \\
$m=1$                   & $23.08 \pm 5.04$            &$25.89 \pm 9.12$           \\
$m=2$                   & $33.86 \pm 7.99$                 & $40.71 \pm 13.56$              \\
$m=\infty$              & $25.53  \pm 8.54$           & $31.02  \pm 11.83$      \\ \hline    
\end{tabular}}
\end{table}

When $m=0$, the routing initialization is uniformly random, yielding the lowest lost demand of 19.66 when $\epsilon = 0.00$. This suggests that random initialization may encourage broad exploration and avoid early overfitting to specific station characteristics. However, its robustness degrades under slight exploration noise, with a notable increase in lost demand to 28.03 when $\epsilon = 0.05$. For $m=1$, the heuristic routing balances distance and inventory level, resulting in lost demands of 23.08 and 25.89 for $\epsilon = 0.00$ and $\epsilon = 0.05$, respectively. This balanced approach provides flexibility and robustness, making it relatively efficient for rebalancing. For $m=2$, the increased sensitivity to both distance and inventory levels leads to a higher lost demand of 33.86, suggesting that excessive prioritization of these factors may cause overfitting and reduce effectiveness in a dynamic environment. Finally, when $m=\infty$, the model relies on a strict greedy heuristic, directing vehicles toward extreme imbalances, but lacks the nuanced approach required for good performance.\\

\begin{figure}[!htbp]
    \centering
    \begin{minipage}{0.48\linewidth}
        \centering
        \includegraphics[width=\linewidth]{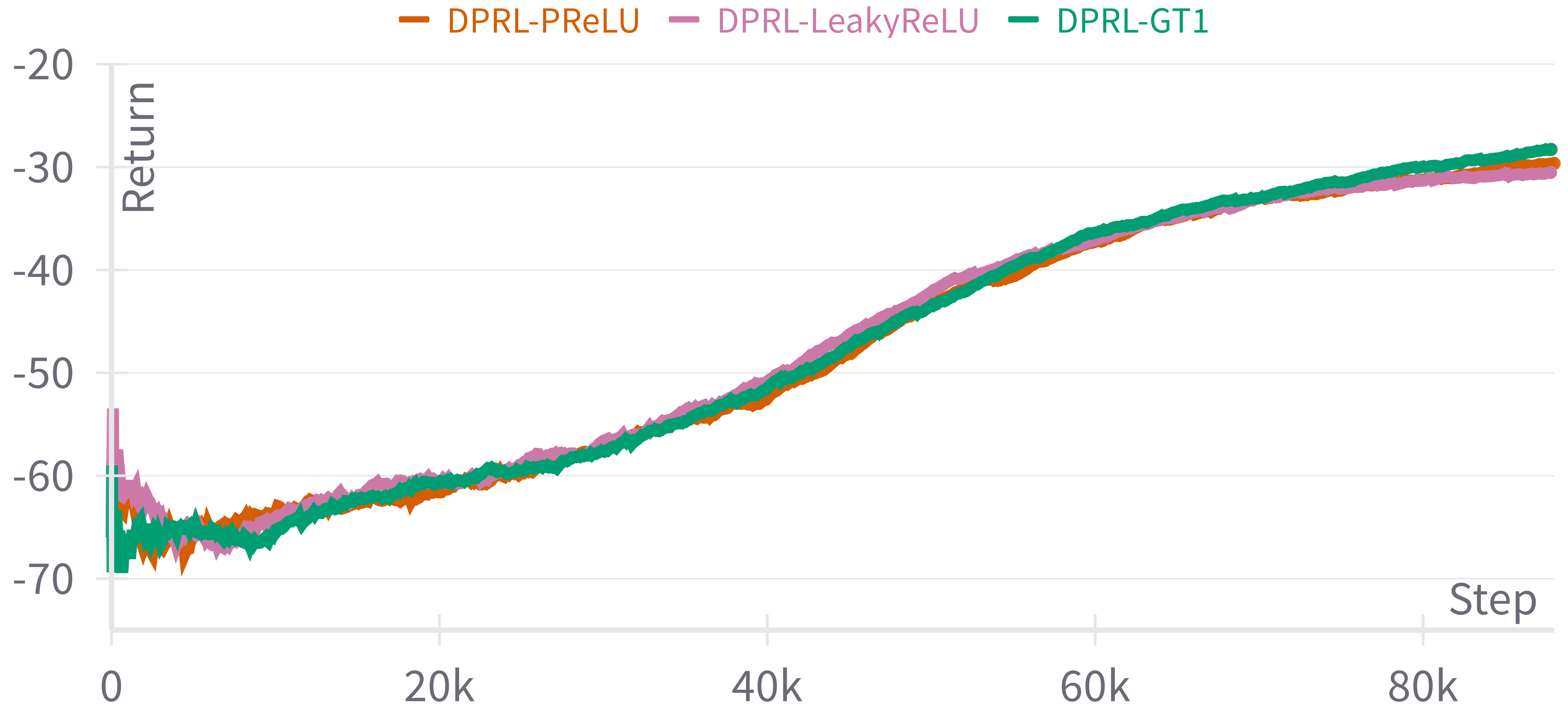}
    \end{minipage}
    \hfill
    \begin{minipage}{0.48\linewidth}
        \centering
        \includegraphics[width=\linewidth]{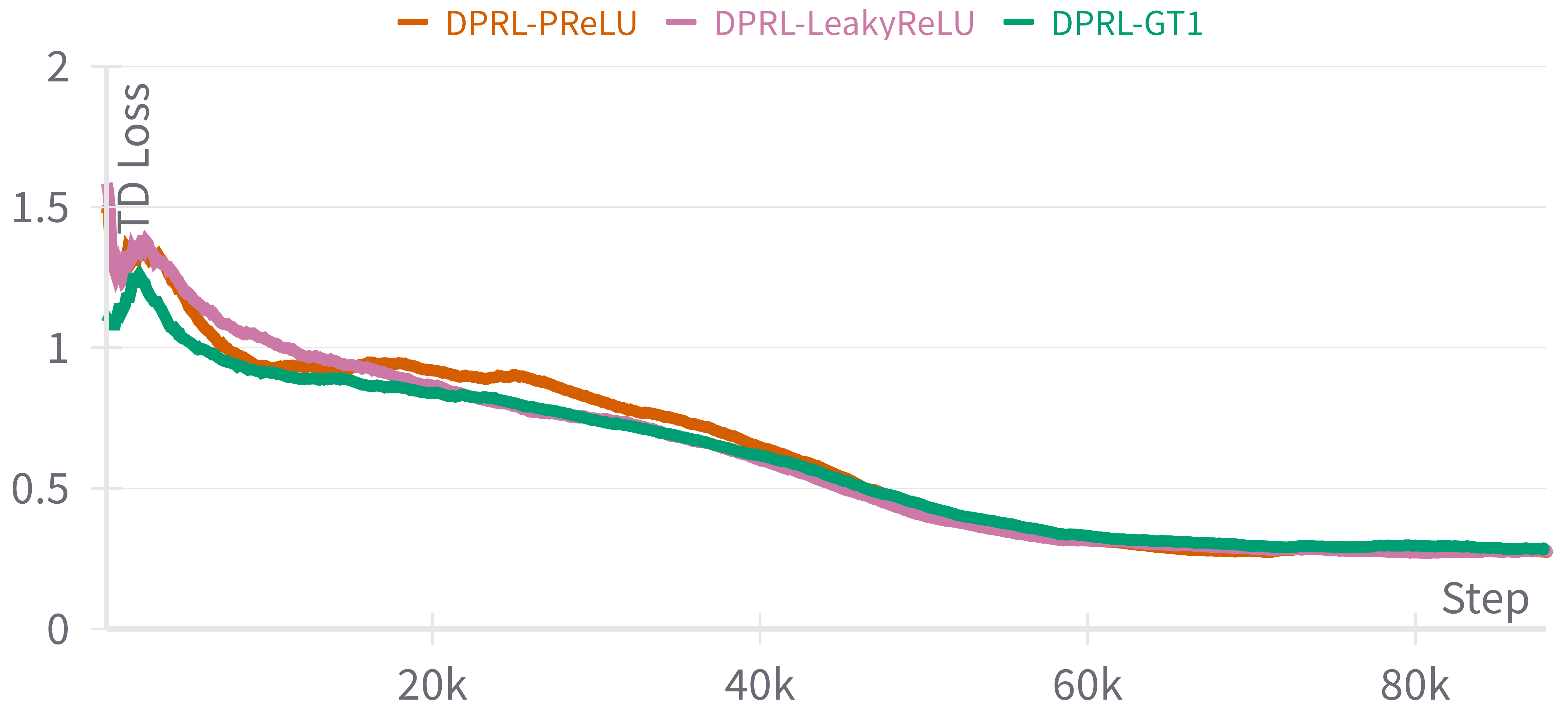}
    \end{minipage}
    \caption{Episodic return and TD loss for DPRL with different activation functions}
    \label{fig:AF}
\end{figure}

\textbf{Policy Architecture.} We further investigate the impacts of different activation functions in the output layer of DPRL. Since the immediate reward in DPRL is defined as the negative value of the lost demand, 
we focus on Leaky ReLU and Parameterized ReLU (PReLU) due to their ability to handle negative outputs \citep{pedamonti2018comparison}. 
Table~\ref{lostdemand_activation} summarizes the average lost demand on the test set under both deterministic ($\epsilon = 0.00$) and slightly stochastic ($\epsilon = 0.05$) evaluation settings. The corresponding training curves are shown in Figure~\ref{fig:AF}.

\begin{table}[!tbhp]
\centering
\caption{Total average lost demand on test set of DPRL with different activation functions}
\label{lostdemand_activation}
\scalebox{0.85}{
\begin{tabular}{l|rr}
\hline
Activation Function                                                 & $\epsilon = 0.00$ & $\epsilon = 0.05$ \\ \hline
DPRL-LeakyReLU     &     22.16        &      29.16        \\
DPRL-PReLU     &   21.73          &    27.45       \\DPRL     &   23.08          &    25.89       \\ \hline        
\end{tabular}}
\end{table}

Overall, the performance differences among the three RL models are not substantial. When $\epsilon = 0.00$, both DPRL-LeakyReLU and DPRL-PReLU slightly outperform the original DPRL, reducing the average lost demand compared to the baseline value of 23.08. However, when $\epsilon = 0.05$, the original DPRL achieves the lowest lost demand,  indicating stronger robustness to exploration-induced randomness. These results suggest that while Leaky ReLU and PReLU may offer marginal benefits in stable or deterministic environments, their performance tends to degrade when exploration is introduced. The inherent flexibility of having no activation function might help navigate the state-action space more effectively in highly exploratory scenarios.

According to Figure~\ref{fig:AF}, the episodic return and TD loss during training are very similar for these three RL models, suggesting that the choice of activation function in the output layer does not significantly alter the learning dynamics.

\begin{figure}[!htbp]
    \centering
    \begin{minipage}{0.48\linewidth}
        \centering
        \includegraphics[width=\linewidth]{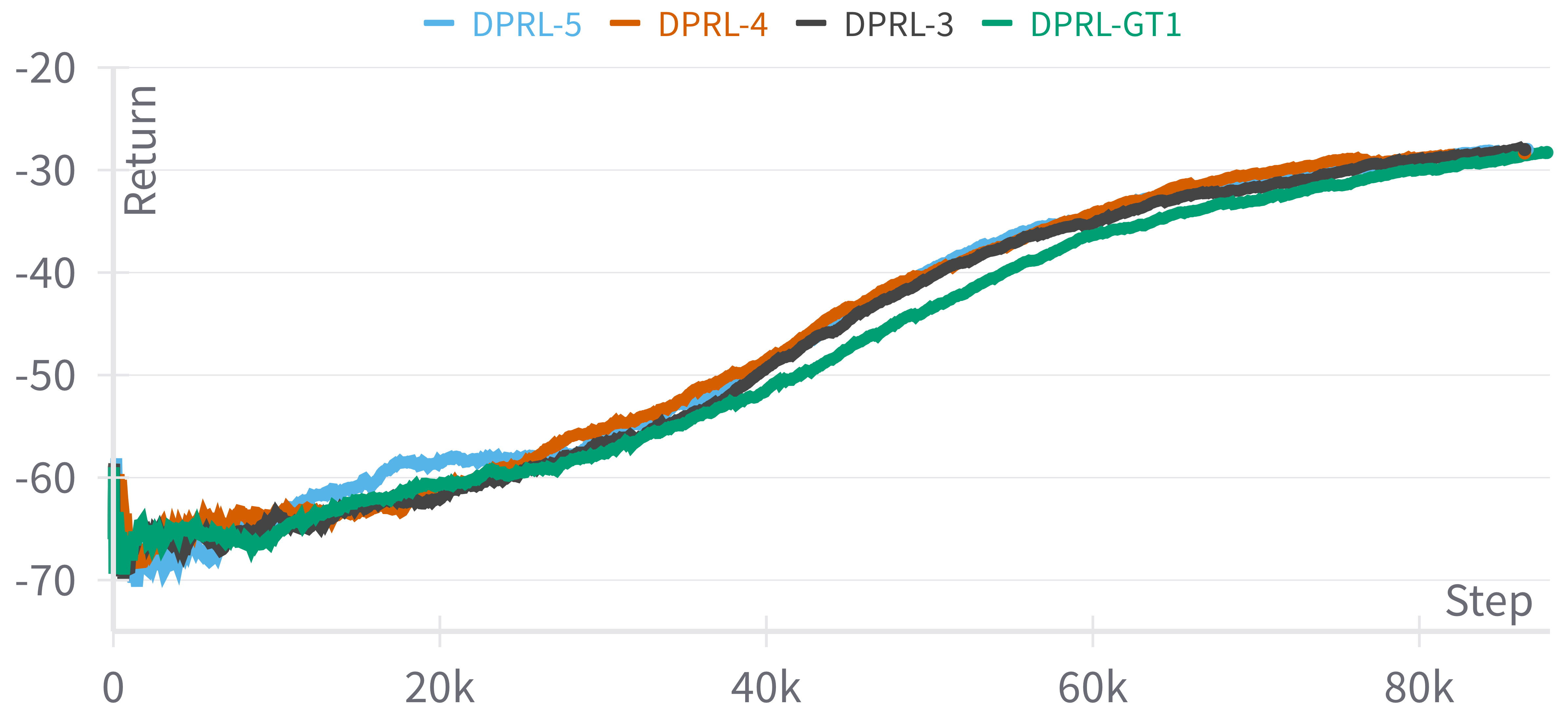}
    \end{minipage}
    \hfill
    \begin{minipage}{0.48\linewidth}
        \centering
        \includegraphics[width=\linewidth]{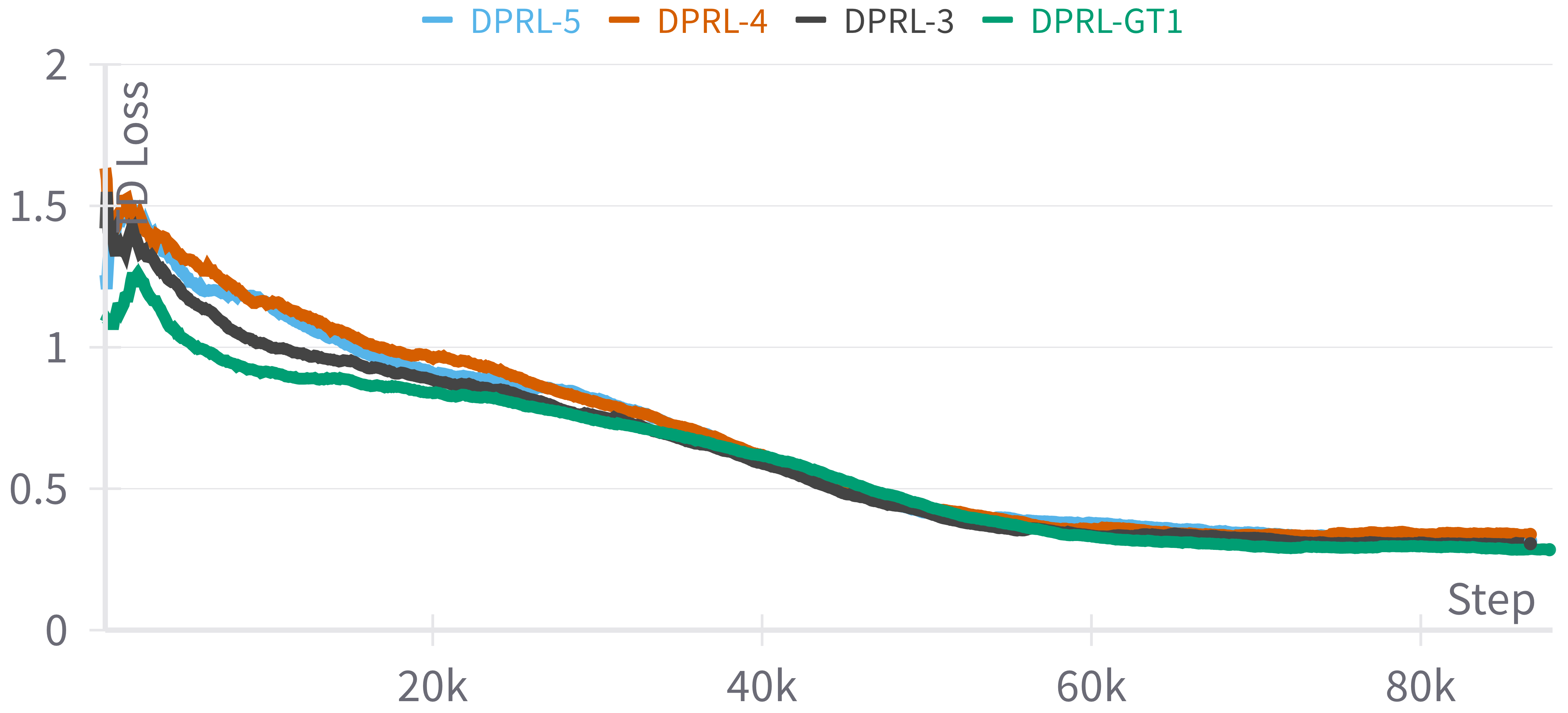}
    \end{minipage}
    \caption{Episodic return and TD loss for DPRL with different layers}
    \label{fig:AL}
\end{figure}

Next, we investigate the impact of network depth by varying the number of fully connected dense layers in the policy networks. As described in Section~\ref{sec:DPRL}, our default DPRL architecture consists of two dense layers following the input layer. In this ablation, we evaluate deeper architectures with three, four, and five layers, denoted as DPRL-3, DPRL-4, and DPRL-5, respectively. The average lost demand on the test set is reported in Table~\ref{lostdemand_al}. Additionally, Figure~\ref{fig:AL} shows the episodic return and TD loss during training.

\begin{table}[!tbhp]
\centering
\caption{Total average lost demand on test set of DPRL with different number of policy network layers}
\label{lostdemand_al}
\scalebox{0.85}{
\begin{tabular}{l|rr}
\hline
\# of Layers                                                 & $\epsilon = 0.00$ & $\epsilon = 0.05$ \\ \hline
DPRL-3     &     19.43       &      26.21        \\
DPRL-4    &   22.03          &    25.45       \\DPRL-5     &   22.25          &    26.70       \\ \hline        
\end{tabular}}
\end{table}

As shown in Table~\ref{lostdemand_al}, increasing the number of layers yields only marginal performance improvements. While deeper architectures enable the model to capture more complex representations and finer-grained patterns in the data, these gains are accompanied by increased model complexity. The larger number of parameters may introduce a greater risk of overfitting. Notably, when $\epsilon = 0.05$, the deeper variants often exhibit reduced performance, likely due to their diminished ability to generalize during exploratory decision-making. These observations highlight a fundamental trade-off between expressiveness and robustness. While deeper models may improve accuracy in stable conditions, they can become brittle when exposed to variability in the environment.

As illustrated in Figure~\ref{fig:AL}, all architecture variants exhibit similar episodic returns, indicating that increasing the network depth does not significantly alter the overall learning process. However, the original DPRL architecture consistently achieves the lowest TD loss, indicating more accurate value estimation.

\section{Reinforcement Learning Basics: Policy Space and Q-value Function.}


In MDPs, a planning solution is encoded through a policy $\pi \in \Pi$, where $\Pi$ denotes the set of all possible policies (policy space). A policy, represented as $\pi(a|s)$, serves as a strategy dictating the probability of taking an action $a$ given state $s$. The overarching objective is to identify the optimal policy $\pi^{*}$ that maximizes the expected cumulative reward over time. 
The performance of a policy can be evaluated using the action-value function (or Q-function), denoted as $Q^{\pi}(\bm{S_{k}},\bm{a_{k}})$, which represents the expected return obtained by taking action $\bm{a_{k}}$ in state $\bm{S_{k}}$ and subsequently following policy $\pi$. This is formalized in Equation \eqref{af}:
\begin{align}
   Q^{\pi}(\bm{S_{k}},\bm{a_{k}}) = \mathbb{E} [R_{k}|\bm{S_{k}},\bm{a_{k}}].
   \label{af}
\end{align}

The goal is to determine the optimal $Q$-function derived to maximize the total expected reward over time, as in Equation (\ref{oaf}). However, directly solving it is often computationally intractable due to the high dimensionality and complexity of the state-action space. 
\begin{align}
   Q^{*}(\bm{S_{k}},\bm{a_{k}}) = \max_{\pi}Q^{\pi}(\bm{S_{k}},\bm{a_{k}}).
   \label{oaf}
\end{align}

In practice, RL algorithms use iterative methods or deep learning-based approximations to estimate the optimal Q-function. Once an approximation is obtained, a greedy policy can be derived by selecting the action with the highest Q-value for any given state. This deterministic policy, defined in Equation~\ref{oaf_ac}, assigns a probability of $1$ to the action that yields the highest Q-value and 0 to all other actions:
\begin{align}
   \pi(\bm{a_{k}}|\bm{S_{k}}) =
\begin{cases} 
1 & \text{if } \bm{a_{k}} = \arg\max_{\bm{a_{k}}} Q^{*}(\bm{S_{k}},\bm{a_{k}}) \\
0 & \text{otherwise}.       \label{oaf_ac}
\end{cases}
\end{align}
This approach simplifies decision-making under uncertainty, guiding actions based on the current knowledge of expected rewards.\\

\end{document}